\documentclass{article}

\usepackage[final,dblblindworkshop]{neurips_2026}
\workshoptitle{Transitioning from Pre-Training to Post-Training}

\usepackage[utf8]{inputenc}
\usepackage[T1]{fontenc}

\usepackage{url}
\usepackage{booktabs}
\usepackage{graphicx}
\usepackage{amsmath}
\usepackage{amssymb}
\usepackage{amsthm}
\usepackage{nicefrac}
\usepackage{microtype}
\usepackage{xcolor}
\usepackage[hidelinks,hyperfootnotes=false]{hyperref}

\newtheorem{proposition}{Proposition}
\newtheorem{lemma}{Lemma}
\newtheorem{corollary}{Corollary}
\theoremstyle{definition}
\newtheorem{definition}{Definition}

\newcommand{\Xset}{\mathcal{X}}
\newcommand{\Yset}{\mathcal{Y}}
\newcommand{\Acc}{\mathcal{A}}
\newcommand{\Hset}{\mathcal{H}}
\newcommand{\Kk}{K_\theta}
\newcommand{\Kcoh}{K_\theta^{\mathrm{coh}}}
\newcommand{\Kxx}{K_{\theta,xx'}}
\newcommand{\pth}{\pi_\theta}
\newcommand{\ueff}{u_{\mathrm{eff}}}
\DeclareMathOperator{\Cov}{Cov}

\newcommand{\CohGold}{+0.03}
\newcommand{\CohHackHi}{+0.29}
\newcommand{\CohHackHiFine}{+0.287}
\newcommand{\CohHackLo}{+0.02}
\newcommand{\CohHackLoFine}{+0.017}
\newcommand{\CopiesHiAcc}{0.959}
\newcommand{\CopiesHiDR}{-0.090}

\newcommand{\CopiesPairedDLHi}{+2.11}
\newcommand{\CopiesPairedDLHiCI}{[+0.75, +3.85]}
\newcommand{\CopiesPairedDLLo}{-1.00}
\newcommand{\CopiesPairedDLLoCI}{[-1.56, -0.36]}
\newcommand{\CopiesPairedDiff}{+0.0191}
\newcommand{\CopiesPairedDiffCI}{[+0.0082, +0.0330]}
\newcommand{\CopiesPairedN}{10}

\newcommand{\CopiesSlope}{+0.0019}
\newcommand{\CopiesSlopeCI}{[-0.0008, +0.0050]}

\newcommand{\KernelRatioNeural}{0.098}
\newcommand{\KernelRatioTab}{exactly $0$}

\newcommand{\LrAdamBDRHi}{+0.028}

\newcommand{\LrAdamCAccLo}{0.930}
\newcommand{\LrAdamCDRHi}{+0.600}
\newcommand{\LrAdamCDRHiSem}{0.084}
\newcommand{\LrAdamCDRLo}{-0.098}

\newcommand{\LrAdamCN}{10}

\newcommand{\LrAdamCNPosHi}{10}

\newcommand{\LrAdamDAccLo}{0.275}

\newcommand{\LrAdamNSat}{Three}

\newcommand{\LrNPosCI}{10}
\newcommand{\LrNTotal}{11}

\newcommand{\MaskFprBaseHi}{0.0149}

\newcommand{\MaskFprBaseLo}{0.0098}

\newcommand{\MaskFprNBelowOne}{6}

\newcommand{\MaskFprPairedLo}{-0.049}
\newcommand{\MaskFprPairedLoCI}{[-0.057, -0.038]}
\newcommand{\MaskFprPairedMid}{-0.069}
\newcommand{\MaskFprPairedMidCI}{[-0.079, -0.060]}

\newcommand{\MaskFprThreeHiDR}{+0.439}

\newcommand{\MaskMassPairedMid}{-0.056}
\newcommand{\MaskMassPairedMidCI}{[-0.064, -0.048]}

\newcommand{\MaskMassThreeHiDR}{+0.427}

\newcommand{\MechCrossHi}{12.2}
\newcommand{\MechCrossHiSem}{0.8}
\newcommand{\MechCrossLo}{6.6}
\newcommand{\MechCrossLoSem}{1.1}
\newcommand{\MechTabCross}{exactly $0$}

\newcommand{\NEnvN}{64}
\newcommand{\NEnvY}{100}
\newcommand{\NEps}{0.03}
\newcommand{\NGoldU}{0.037}
\newcommand{\NGrpoG}{8}

\newcommand{\NParams}{101,706}
\newcommand{\NSteps}{400}
\newcommand{\NThreshValidity}{18}
\newcommand{\NThreshValidityLo}{1.1}
\newcommand{\NuFiveACrossU}{$u=0.27$}
\newcommand{\NuFiveADHi}{+0.663}
\newcommand{\NuFiveADHiSem}{0.079}
\newcommand{\NuFiveADLo}{-0.103}
\newcommand{\NuFiveADLoSem}{0.004}

\newcommand{\NuFiveANSeeds}{10}

\newcommand{\NuFiveBAccKLLo}{0.821}

\newcommand{\NuFiveBAccNoKLLo}{0.978}
\newcommand{\NuFiveBBeta}{0.3}

\newcommand{\NuFiveBDRKLLo}{0.000}

\newcommand{\NuFiveBDRNoKLLo}{-0.104}

\newcommand{\NuFiveBNSeeds}{10}
\newcommand{\NuFiveCAccI}{0.909}

\newcommand{\NuFiveCAccU}{0.987}

\newcommand{\NuFiveCGap}{8}
\newcommand{\NuFiveCGapCI}{[4, 13]}
\newcommand{\NuFiveCGapHiLr}{38}
\newcommand{\NuFiveCGapHiLrCI}{[22, 55]}
\newcommand{\NuFiveCGapHiLrNPos}{9}
\newcommand{\NuFiveCGapHiLrSem}{9}

\newcommand{\NuFiveCGapRL}{2}
\newcommand{\NuFiveCGapRLCI}{[-3, 7]}
\newcommand{\NuFiveCGapRLHiLr}{32}
\newcommand{\NuFiveCGapRLHiLrCI}{[16, 48]}
\newcommand{\NuFiveCGapSem}{2}
\newcommand{\NuFiveCGapTab}{6}
\newcommand{\NuFiveCGapTabCI}{[5, 8]}

\newcommand{\NuFiveCLkyMin}{0.973}

\newcommand{\NuFiveCNSeeds}{10}
\newcommand{\NuFourMatchedGap}{0.045}
\newcommand{\NuFourMatchedGapLo}{0.016}

\newcommand{\NuFourNSeeds}{10}
\newcommand{\NuFourNShared}{5}

\newcommand{\NuFourSameM}{16}
\newcommand{\NuFourSameMBig}{-0.081}
\newcommand{\NuFourUOneBig}{+0.021}
\newcommand{\NuFourUOneBigSem}{0.028}
\newcommand{\NuFourUOneSmall}{-0.024}
\newcommand{\NuFourUOneSmallSem}{0.053}

\newcommand{\NuSixBaseSpread}{0.0103}

\newcommand{\NuSixSlope}{+0.0270}

\newcommand{\NuSixTabSlope}{-0.0001}
\newcommand{\NuThreeBaseRhoLoK}{0.129}
\newcommand{\NuThreeBaseRhoSpreadK}{0.080}
\newcommand{\NuThreeDLHiK}{-6.85}
\newcommand{\NuThreeDLHiU}{-4.15}
\newcommand{\NuThreeDLLoK}{-5.54}
\newcommand{\NuThreeDLLoU}{-7.19}
\newcommand{\NuThreeEtaK}{12\%}
\newcommand{\NuThreeEtaM}{43\%}
\newcommand{\NuThreeFprRange}{1.0--8.1\%}
\newcommand{\NuThreeKRange}{1--8}

\newcommand{\NuThreeNSeeds}{10}
\newcommand{\NuTwoBaseAcc}{0.441}
\newcommand{\NuTwoBaseLky}{0.475}
\newcommand{\NuTwoBaseRhoMean}{0.128}
\newcommand{\NuTwoBaseRhoSpread}{0.015}

\newcommand{\NuTwoFlipHi}{16}

\newcommand{\NuTwoFlipLo}{2}
\newcommand{\NuTwoHiSeedList}{0.02, 0.06, 0.06, 0.08, 0.08, 0.14, 0.17, 0.20, 0.22, 0.56}
\newcommand{\NuTwoHiSeedMedian}{0.11}

\newcommand{\NuTwoNSeeds}{10}
\newcommand{\NuTwoNeuralHiAcc}{0.840}
\newcommand{\NuTwoNeuralHiDL}{-3.7}
\newcommand{\NuTwoNeuralHiDR}{+0.028}
\newcommand{\NuTwoNeuralHiDRSem}{0.050}
\newcommand{\NuTwoNeuralHiRho}{0.160}
\newcommand{\NuTwoNeuralHiRhoSem}{0.050}
\newcommand{\NuTwoNeuralLoAcc}{0.974}
\newcommand{\NuTwoNeuralLoDL}{-9.0}
\newcommand{\NuTwoNeuralLoDR}{-0.104}
\newcommand{\NuTwoNeuralLoDRSem}{0.004}
\newcommand{\NuTwoNeuralLoRho}{0.025}
\newcommand{\NuTwoNeuralLoRhoSem}{0.005}
\newcommand{\NuTwoNeuralSlope}{+0.0177}
\newcommand{\NuTwoSgdCrossU}{$u=0.07$}

\newcommand{\NuTwoSgdNeuralHiAcc}{0.000}
\newcommand{\NuTwoSgdNeuralHiRho}{1.000}

\newcommand{\NuTwoSgdNeuralLoAcc}{0.926}
\newcommand{\NuTwoSgdNeuralLoRho}{0.051}

\newcommand{\NuTwoTabSlope}{+0.0028}
\newcommand{\NullBetaRange}{$[0.05, 2.0]$}
\newcommand{\NullDriftKL}{$8.6\times 10^{-9}$}
\newcommand{\NullDriftMirror}{$8.6\times 10^{-9}$}
\newcommand{\NullDriftMirrorLam}{$9.6\times 10^{-8}$}
\newcommand{\NullDriftMirrorPooled}{+0.057}
\newcommand{\NullMirrorSteps}{400}
\newcommand{\ProbeCap}{8}
\newcommand{\ProbeNCond}{28}
\newcommand{\ProbeNPts}{672}
\newcommand{\ProbeNSeeds}{3}
\newcommand{\ProbeNegKFrac}{13\%}
\newcommand{\ProbeNegWeightShare}{9\%}
\newcommand{\ProbeResidShareMedian}{0.9}
\newcommand{\ProbeSignAgreeCond}{93\%}
\newcommand{\ProbeSignAgreeLeadCond}{71\%}
\newcommand{\ProbeSignAgreeLeadPts}{53\%}
\newcommand{\ProbeSignAgreePts}{70\%}
\newcommand{\ProbeSteps}{0, 3, 6, 10, 15, 20, 40}
\newcommand{\ProbeVbarHiLast}{0.994}
\newcommand{\ProbeVbarLoLast}{0.848}
\newcommand{\ProbeVbarStepLast}{40}
\newcommand{\RepENineCoh}{+0.255}

\newcommand{\RepENineLoCoh}{+0.240}
\newcommand{\RepENinePh}{0.0151}
\newcommand{\RepEOneCoh}{+0.294}

\newcommand{\RepEOneLoCoh}{+0.090}
\newcommand{\RepEOneLoPh}{0.0328}
\newcommand{\RepEOneLoPhOut}{0.0054}
\newcommand{\RepEOnePh}{0.0355}
\newcommand{\SharedSlopeMatched}{+0.0210}

\newcommand{\UeffENineFour}{0.19}
\newcommand{\UeffENineHi}{1.00}
\newcommand{\UeffENineLo}{0.08}
\newcommand{\UeffENineSixteen}{0.33}
\newcommand{\UeffEOneFour}{0.64}
\newcommand{\UeffEOneHi}{1.00}
\newcommand{\UeffEOneLo}{0.65}
\newcommand{\UeffEOneSixteen}{0.94}
\newcommand{\UeffMassENineFour}{0.06}
\newcommand{\UeffMassENineHi}{1.00}
\newcommand{\UeffMassENineLo}{0.01}
\newcommand{\UeffMassENineSixteen}{0.25}
\newcommand{\UeffMassEOneFour}{0.23}
\newcommand{\UeffMassEOneHi}{1.00}
\newcommand{\UeffMassEOneLo}{0.09}
\newcommand{\UeffMassEOneSixteen}{0.68}

\AddToHook{para/begin}[loose]{\looseness=-1}

\title{Wrong the Same Way: Cross-Prompt Sharing Both\\Suppresses and Amplifies Reward Hacking in RLVR}

\author{%
  Michael Li\thanks{Equal contribution. Our source code is available at \href{https://github.com/Rohan-Pandey1729/RLVR-Reward-Hacking}{\nolinkurl{github.com/Rohan-Pandey1729/RLVR-Reward-Hacking}}.} \\
  Carnegie Mellon University \\
  \texttt{ml7@andrew.cmu.edu} \\
  \And
  Rohan Pandey\footnotemark[1] \\
  University of Washington \\
  \texttt{rpande@uw.edu} \\
  \And
  Ray Chen\footnotemark[1] \\
  Independent \\
  \texttt{raychen256@gmail.com} \\
}

\begin{document}

\maketitle

\begin{abstract}
\looseness=-1
Two accounts of how reinforcement learning with verifiable rewards (RLVR) interacts with an imperfect verifier coexist: in one, reward hacking is inherited from the base model; in the other, RL amplifies it. We study both in a controlled environment small enough to compute the entire response distribution exactly. We vary the universality of a leak, or the fraction of prompts that accept the same wrong response, and the verifier's static false-positive rate. At its KL-regularized optimum, RLVR acts as a filter that reweights the accepted set without altering its composition, so the hack rate is inherited exactly. Taking that as a null model, a known decomposition writes the drift from it as a sum over prompts of covariances between reward and the empirical tangent kernel. The terms from other prompts vanish without parameter sharing. With sharing, every prompt that accepts a wrong answer pushes it up and every prompt that rejects it pushes it down through the baseline, in proportion to the model's probability of producing it there. A shared-parameter policy thus drives a leak that few prompts accept below the base rate and amplifies one that most prompts accept. At every convergent learning rate under Adam, SGD, and sampled GRPO, hacking ends higher at high universality than at low, but whether either end crosses the base rate depends on the step size. Independent per-prompt copies remove the dependence. The direction of this push depends on neither the verifier's error rate nor universality alone, but on how much of the model's kernel-weighted probability of a wrong answer falls on prompts that accept it, which frozen base model controls support. Verifier fixes should target the mistakes the model repeats across prompts: two patches that remove equally many errors differ by up to \NuFiveCGapHiLr{} points of true accuracy. We propose this alignment as a hypothesis for why earlier studies disagree.
\end{abstract}

\section{Introduction}

In reinforcement learning with verifiable rewards (RLVR), a response is rewarded when a programmatic verifier accepts it \citep{shao2024deepseekmath, deepseekai2025r1}. Because verifiers are imperfect proxies for correctness, policies are vulnerable to \emph{reward hacking}: optimizing for the verifier's acceptance criteria rather than the intended task objective \citep{skalse2022defining}. Generative judges accept non-word symbols and generic reasoning openers \citep{zhao2025onetoken}, test suites accept programs that pass only the visible tests \citep{zhang2026leaky}, and outcome verifiers reward a correct final answer reached by invalid reasoning \citep{yuan2026miracle}, one reason process supervision \citep{lightman2024verify} was proposed over outcome supervision \citep{cobbe2021gsm8k}. Every such false positive provides an exploitable shortcut where the policy is rewarded without being correct.

Two accounts of how RL interacts with these false positives coexist. In the first, RLVR is robust to them and any hacking is inherited from the base model: verifier noise up to $20\%$ leaves validation reward within one point of clean training \citep{plesner2026imperfect}, noise rescales convergence time rather than the destination while the verifier's Youden index stays positive \citep{rad2026rateorfate}, and a preregistered comparison of leaky and hardened test suites suggests that the wrong programs a trained model emits were already emitted by the base model \citep{zhang2026leaky}. In the second, RLVR is vulnerable to them and actively amplifies hacking: RL amplifies hacking seeded by a trace of contamination \citep{khalifa2026countdown}, systematic verifier error produces delay, plateau, or collapse \citep{egashira2026delay}, models learn to game extensional verifiers \citep{helff2026gaming}, and production reward hacking generalizes into broader misalignment \citep{macdiarmid2025natural}. The studies behind these accounts use diverse setups, so they do not identify what decides the outcome.

Theory sides with the first account. At the KL-regularized optimum of the RLHF objective \citep{stiennon2020summarize, ouyang2022instructgpt} the policy is $\pi^\star(y \mid x) \propto \pi_{\mathrm{ref}}(y\mid x)\,e^{r(x,y)/\beta}$, so within any reward level set the odds between responses are those of the reference model \citep{korbak2022bayesian}. RLVR thus acts as a filter on the accepted set: it changes how much mass the set receives, not its composition, which is consistent with RLVR sharpening rather than expanding a base model's abilities \citep{yue2025rlincentivize, gai2025smoothing, kruszewski2026filtering, dymetman2026binary}. For verifier false positives, this predicts inheritance. However, training in practice takes first-order steps on a model whose parameters are shared across prompts, often with little or no KL term; we ask what happens to verifier false positives under such training.

\begin{figure}[t]
\centering
\includegraphics[width=\linewidth]{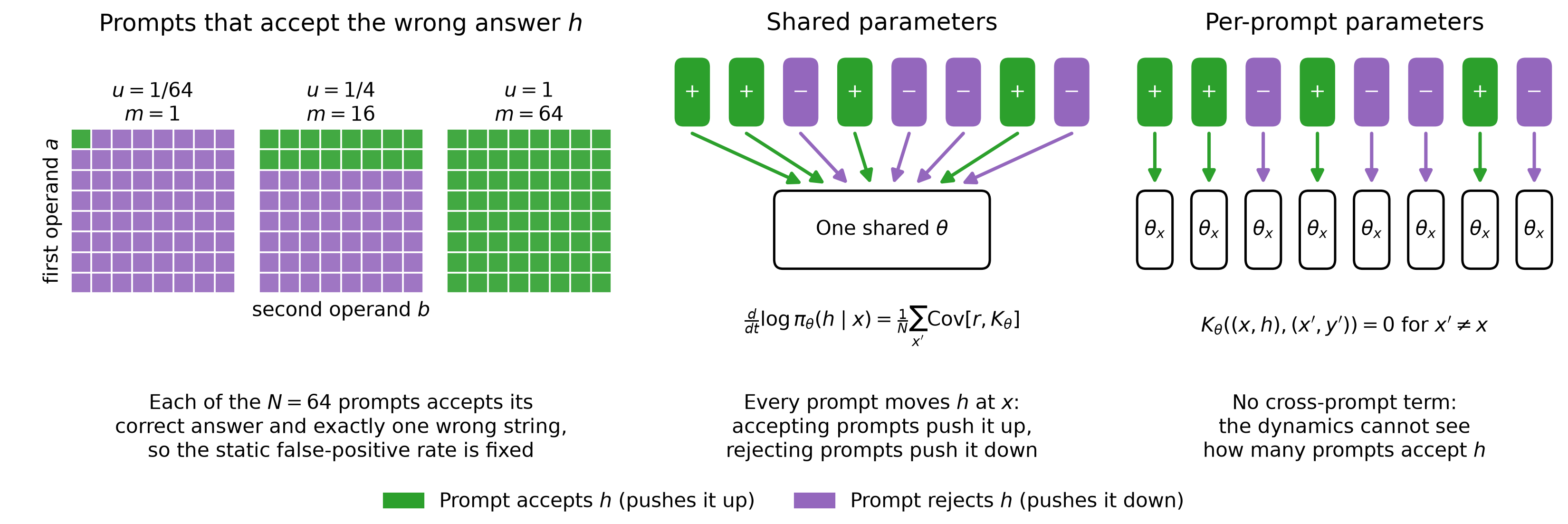}
\caption{The environment and the mechanism. Left: the $N=64$ prompts, colored by whether the verifier accepts a designated wrong answer $h$; every prompt accepts exactly one wrong string, so sweeping $u=m/N$ leaves the static false-positive rate unchanged. Center and right: with shared parameters every prompt's gradient lands on the same weights, so $h$ is pushed up by the prompts that accept it and down by those that reject it (Lemma~\ref{lem:drift}); with per-prompt parameters the cross-prompt term is identically zero (Corollary~\ref{cor:tabular}).}
\label{fig:overview}
\end{figure}

We answer the question in a controlled environment (Figure~\ref{fig:overview}): single-digit multiplication with a two-digit answer space, in which the full response distribution of a small transformer is enumerated exactly at every step. The verifier accepts the correct answer plus designated wrong strings, and we vary how many prompts share the same wrong string, its \emph{universality}, at a fixed false-positive rate. Our contributions are as follows:

\begin{itemize}\itemsep1pt
\item \textbf{We identify a two-sided effect of parameter sharing.} Sharing suppresses hacks shared among few prompts and amplifies hacks shared among many prompts. Hacking rises with universality at every convergent learning rate under three training setups, though the sign of the drift depends on the step size, and per-prompt copies of the model remove the effect (Sections~\ref{sec:lr} and~\ref{sec:mechanism}).
\item \textbf{We identify what sets the direction.} By a known decomposition, other prompts affect a hack only through shared parameters. With aligned gradients, the net update to that string pushes it up when the kernel-weighted share of its probability on accepting prompts exceeds the model's pass rate. Frozen base model controls support this (Sections~\ref{sec:theory} and~\ref{sec:controls}).
\item \textbf{We identify what to audit.} Two verifier patches that remove equally many errors differ by up to $\NuFiveCGapHiLr{}$ points of true accuracy over \NuFiveCNSeeds{} seeds (Section~\ref{sec:hardening}). Audits should check whether accepted wrong answers are ones the model already produces across prompts.
\end{itemize}

\section{Background}

\citet{egashira2026delay} established the premise we build on: whether training plateaus or collapses depends on the ``specific pattern of introduced errors'' rather than on the blanket error rate. The decomposition in Section~\ref{sec:theory} follows \citet{ren2025dynamics} for fine-tuning and \citet{tomihari2026rldynamics} for RL post-training. We do not claim the decomposition; what we use in our work is the push-down: a rewarded response on one prompt is penalized through the baseline on every prompt that rejects it, so the mechanism that amplifies a widely accepted wrong answer suppresses a narrowly accepted one. Appendix~\ref{app:related} covers further background and related work.

\section{Setup: RLVR with an Exploitable Verifier}
\label{sec:setup}

Let $\Xset$ be a finite prompt set with $|\Xset| = N$ and where each prompt is equally weighted, and let $\Yset$ be a finite response space. A response is correct if $c(x,y)=1$; we refer to the correct response as $g_x$, which is unique in our setting. The deployed verifier returns $r(x,y)\in\{0,1\}$ with $r \geq c$, so it has no false negatives. $\Acc(x)=\{y: r(x,y)=1\}$ is the accepted set and $\Hset(x) = \{y : r(x,y)=1,\, c(x,y)=0\}$ its false-positive part, or the \emph{leaks}. RLVR maximizes $J(\theta) = \mathbb{E}_{x}\,\mathbb{E}_{y\sim\pth(\cdot\mid x)}[r(x,y)]$.

\begin{definition}[Hack fraction]
The hack fraction is $\rho(\pi) = \frac{1}{N}\sum_{x} \pi(\Hset(x)\mid x)\,/\,\pi(\Acc(x)\mid x)$, the probability that a rewarded response is a false positive, averaged over prompts. We also track the prompt-averaged log-odds $\Lambda(\pi) = \frac1N \sum_x \log \big[\pi(\Hset(x)\mid x)\,/\,\pi(g_x\mid x)\big]$. Both are finite under any softmax parameterization if no $\Hset(x)$ is empty.
\end{definition}

\begin{definition}[Leak universality]
\label{def:univ}
For a response $h$, let leak count $m(h) = |\{x : h \in \Hset(x)\}|$ and leak universality $u(h) = m(h)/N$. A leak with $u=1$ is accepted on every prompt, which we call \emph{universal}; a leak with $u = 1/N$ is accepted on one prompt only, which we call \emph{idiosyncratic}.
\end{definition}

\subsection{The filter baseline}

\begin{proposition}[RLVR is a filter; known]
\label{prop:filter}
Assume $r \geq c$ and let $\pi_{\mathrm{ref}}(\cdot\mid x)$ have full support. Let $\pi^\star_\beta = \arg\max_\pi J(\pi) - \beta\,\mathbb{E}_x \mathrm{KL}(\pi(\cdot\mid x)\Vert \pi_{\mathrm{ref}}(\cdot\mid x))$. Then $\pi^\star_\beta(y\mid x) \propto \pi_{\mathrm{ref}}(y\mid x)\, e^{r(x,y)/\beta}$, so for any $y,y'$ with $r(x,y)=r(x,y')$ the odds are unchanged. Consequently $\rho(\pi^\star_\beta) = \rho(\pi_{\mathrm{ref}})$ and $\Lambda(\pi^\star_\beta) = \Lambda(\pi_{\mathrm{ref}})$ for every $\beta>0$. The same invariance holds at every step of any multiplicative update $\pi_{t+1}(\cdot\mid x)\propto \pi_t(\cdot\mid x)\exp(\eta f(r(x,\cdot)))$.
\end{proposition}

The closed form is the Gibbs variational principle, standard in RL \citep{peters2010reps} and in the LLM setting \citep{korbak2022bayesian, kruszewski2026filtering, dymetman2026binary}, and it is the baseline every departure below is measured against. As $\beta\to0$, $\pi^\star_\beta \to \pi_{\mathrm{ref}}(\cdot\mid\Acc(x))$: along the whole regularization path the accepted set keeps the reference model's composition, so nothing inside it can move. The pooled share $\sum_x \pi(\Hset(x)\mid x)/\sum_x \pi(\Acc(x)\mid x)$ that prior work reports is a weighted average of these invariant per-prompt ratios with weights that RLVR moves, so it drifts under pure filtering (by \NullDriftMirrorPooled{} in our environment; Appendix~\ref{app:proofs}); we report per-prompt statistics throughout.

\section{Why First-Order Training Departs From Filter}
\label{sec:theory}

In practice, RLVR takes first-order steps on a shared-parameter model, often with a small or absent KL term. The following decomposition, following \citet{ren2025dynamics} for fine-tuning and \citet{tomihari2026rldynamics} for RL post-training, describes the resulting drift exactly.

\begin{lemma}[Drift decomposition; known]
\label{lem:drift}
Under exact-gradient flow $\dot\theta = \nabla_\theta J(\theta)$,
\begin{equation}
\frac{d}{dt}\log\pth(y\mid x)
= \frac{1}{N}\sum_{x' \in \Xset}
\Cov_{y'\sim\pth(\cdot\mid x')}\!\Big[\, r(x',y'),\; \Kk\big((x,y),(x',y')\big)\Big],
\label{eq:drift}
\end{equation}
where $\Kk((x,y),(x',y')) = \langle \nabla_\theta\log\pth(y\mid x), \nabla_\theta\log\pth(y'\mid x')\rangle$ is the empirical tangent kernel of the log-policy (Appendix~\ref{app:proofs}).
\end{lemma}

The $x'=x$ term is the ordinary within-prompt (own-prompt) update, present for any policy; the $x'\neq x$ (cross-prompt) terms are the contribution of parameter sharing. Most of our experiments use Adam \citep{kingma2015adam}, whose momentum makes the update history-dependent, so the identity does not apply verbatim; Corollary~\ref{cor:tabular} below still holds, because Adam's state is per-coordinate, and we also report plain SGD. The kernel is the finite-width empirical kernel, not the infinite-width limit of \citet{jacot2018ntk}.

\begin{corollary}[Per-prompt parameters cannot see universality]
\label{cor:tabular}
If $\theta = (\theta_x)_{x\in\Xset}$ and $\pth(\cdot\mid x)$ depends on $\theta_x$ only, then $\Kk((x,y),(x',y'))=0$ for $x'\neq x$, \eqref{eq:drift} collapses to its $x'=x$ term, and the trajectory of each block $\theta_x$ is an autonomous ODE that depends on the verifier only through $r(x,\cdot)$. Such a policy can depend on $m$ only through what it inherited at initialization.
\end{corollary}

A per-prompt policy is therefore a control rather than a baseline. It does not sit \emph{on} the filter baseline: for two equally rewarded responses $y_1,y_2$ at a prompt $x$, gradient flow gives $\tfrac{d}{dt}\log(\pi_1/\pi_2) = \tfrac1N(\pi_1-\pi_2)(1-V(x))$ with $\pi_i=\pth(y_i\mid x)$ and pass rate $V(x)=\sum_y\pth(y\mid x)\,r(x,y)$, the rich-get-richer behavior of softmax policy gradient \citep{mei2020global, lochab2026ucpo}. That drift is prompt-local and blind to $u$, which is what a control requires.

\paragraph{The quantity that matters.} Equation~\eqref{eq:drift} does not isolate $u=m/N$. Specializing it to a leak $h$ accepted on a set $U$ of prompts, the acceptance fraction that appears is kernel- and policy-weighted:
\begin{equation}
\ueff(x,h) \;=\; \Big(\textstyle\sum_{x' \in U} w_{x'}\Big) \Big/ \Big(\textstyle\sum_{x'} w_{x'}\Big),
\qquad w_{x'} = \Kxx\,\pth(h\mid x'),
\label{eq:ueff}
\end{equation}
where $\Kxx = \Kk((x,h),(x',h))$ couples $h$ at $x$ to the same string at $x'$. The weights can be negative in general (\ProbeNegKFrac{} of the entries are at our base models; Appendix~\ref{app:probe}); when they are positive, $\ueff$ is the share of the model's kernel-weighted mass on $h$ that sits on prompts which accept $h$. The raw fraction $m/N$ is recovered only when the kernel entries and the probabilities are the same at every prompt.

\begin{definition}[Coherent kernel model]
\label{def:coherent}
Write $\Kk = \Kcoh + R$ with $\Kcoh((x,h),(x',y')) = \kappa_{xx'}\,\mathbf{1}[y'=h] + a(x,x') + b(x,x',y')$, where $a$ does not depend on $y'$ and $b$ does not depend on $h$. The $a$ and $b$ terms drop out of \eqref{eq:drift} and of the within-prompt log-odds respectively (Appendix~\ref{app:proofs}), leaving the same-string entries $\kappa_{xx'}$ and a residual $R$ that we do not bound. Coherence is a property of how a response is computed, not of string identity: a wrong answer produced by prompt-independent machinery is coherent, while correct answers that coincide on two prompts need not be.
\end{definition}

\begin{proposition}[Universality threshold]
\label{prop:threshold}
Let $h$ be accepted on $U$, let the correct answer be incoherent across prompts (its same-string entries $\kappa^{g}_{xx'}$ vanish for $x'\neq x$) with self-scale $\kappa_{s,g}=\kappa^{g}_{xx}$, and let the pass rate be the same on every prompt, $V(\cdot)\equiv\bar V$, which makes $\bar V=J(\theta)$ the policy's verifier pass rate. Then at any prompt $x$ the coherent part of the kernel contributes
\begin{equation}
\frac{d}{dt}\Big[\log\pth(h\mid x) - \log\pth(g_x\mid x)\Big]\Big|_{\mathrm{coh}}
= \frac{W_x}{N}\,\big(\ueff(x,h)-\bar V\big) \;-\; \frac{1-\bar V}{N}\,\kappa_{s,g}\,\pth(g_x\mid x),
\label{eq:threshold}
\end{equation}
with $W_x=\sum_{x'}\kappa_{xx'}\,\pth(h\mid x')$ and $\ueff$ taken with $\kappa_{xx'}$ in place of $\Kxx$. If in addition $x\in U$, $\pth(h\mid\cdot)\equiv p_h$, and $\kappa_{xx'}$ equals $\kappa_s$ on the diagonal and $\kappa>0$ off it, then with $p_g=\pth(g_x\mid x)$ this is $\kappa\,p_h\,(u-\bar V) + \frac{1-\bar V}{N}\big[(\kappa_s-\kappa)\,p_h - \kappa_{s,g}\,p_g\big]$, whose leading-order sign is that of $u-\bar V$ (Appendix~\ref{app:proofs}).
\end{proposition}

Three consequences follow. The verifier's false-positive rate does not appear: accepting more distinct wrong responses per prompt changes every static (policy-independent) audit metric and leaves \eqref{eq:threshold} unchanged. Sharing is two-sided: every prompt that rejects a wrong answer contributes a negative advantage, so the term that amplifies a widely accepted wrong answer pushes a narrowly accepted one down, and a per-prompt policy has no such term. And the threshold rises during training with $\bar V$, so a wrong answer can be amplified early and suppressed later. Proposition~\ref{prop:threshold} describes one channel of the drift, the same-string channel, at leading order under homogeneity assumptions that our base models violate; it does not predict where the final drift crosses zero, and raw $u$ is not the controlling variable in our headline sweep (Figure~\ref{fig:universality}), where the base model's supervised training concentrates a leak's mass on the prompts that accept it (Section~\ref{sec:controls}; Appendix~\ref{app:probe} compares the channel with the exact cross-prompt drift).

\section{Experiments}
\label{sec:design}

\paragraph{Task and verifier.} We use single-digit multiplication as the task. $\Xset$ consists of $N=\NEnvN{}$ pairs $(a,b)$ where $a,b\in\{2,\dots,9\}$. The correct response is the two-digit decimal string of $a\times b$, and the response space is all $\NEnvY{}$ two-digit strings ranging from ``00'' to ``99'', so that $\pth(\cdot\mid x)$ is enumerable for every prompt at every step. Note that there exist response strings that can never be the correct answer to any input $(a,b)$. The verifier accepts the correct string plus $k$ designated wrong strings per prompt, none of which is the correct answer of any prompt; prompts, ordered by $(a,b)$, are partitioned into contiguous blocks of size $m$ whose members share the same leak strings, so varying $m$ sets $u=m/N$ at a fixed static false-positive rate $k/(|\Yset|-1)$, and varying $k$ sweeps the false-positive rate at fixed universality.

\paragraph{Base model.} The policy is a $2$-layer, $64$-dimensional transformer \citep{vaswani2017attention} with \NParams{} parameters, trained from scratch on a supervised target. At prompt $x$, the target gives the correct answer with probability $\lambda_x$ and a uniformly random string otherwise, and a share $\varepsilon=\NEps{}$ of it is replaced by leak demonstrations, since a trace of contamination suffices to seed hacking \citep{khalifa2026countdown}. The $\lambda_x$ are evenly spaced over $[0,0.9]$, so the model solves some prompts most of the time and others almost never, and a fixed permutation assigns them to prompts independently of the blocks. The base model has true accuracy \NuTwoBaseAcc{} (probability of the correct answer) and hack fraction \NuTwoBaseRhoMean{}. Its hack fraction varies by \NuTwoBaseRhoSpread{} across conditions, so we report hacking as a drift from each run's own base model, $\Delta\rho$ for the hack fraction and $\Delta\Lambda$ for the log-odds, which Proposition~\ref{prop:filter} makes exactly zero under the filter baseline (verified to floating-point precision in Appendix~\ref{app:details}).

\paragraph{Per-prompt control.} The transformer (the \emph{shared} arm) is paired with a tabular policy of free per-prompt parameters under the same autoregressive factorization (the \emph{per-prompt} arm), initialized to reproduce the base transformer's distribution to $10^{-4}$; both are trained on the same objective and learning-rate grid (Table~\ref{tab:protocol}).

\paragraph{Protocol.} Unless stated otherwise we run exact-expectation policy gradient with the exact mean baseline, the expectation limit of RLOO \citep{ahmadian2024rloo} and Dr.\ GRPO \citep{liu2025drgrpo}, for \NSteps{} Adam steps at ten seeds per cell over the learning-rate grid $\{3\times10^{-4},10^{-3},3\times10^{-3},10^{-2},3\times10^{-2}\}$, and report the \emph{operating point} a practitioner would select without ground truth: the smallest learning rate whose final verifier reward is within $10^{-3}$ of the best, a conservative tie-break, since smaller steps drift less from the filter baseline.

\section{Results}
\label{sec:results}

\begin{figure}[t]
\centering
\includegraphics[width=\linewidth]{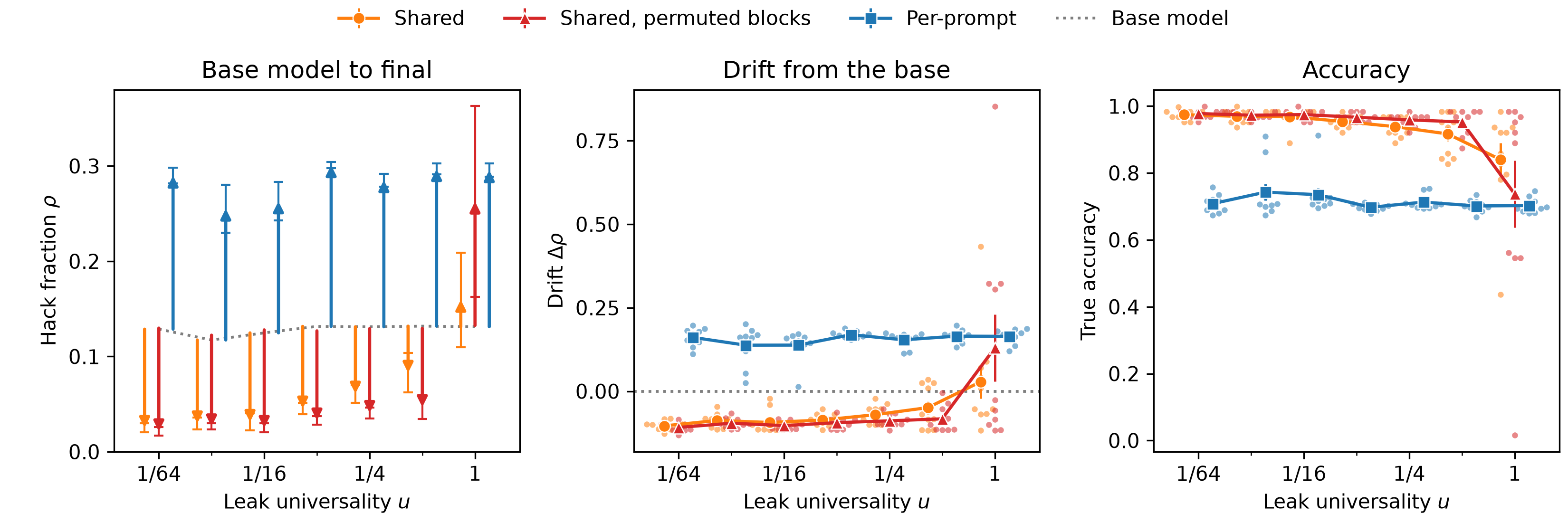}
\caption{At a fixed false-positive rate, the more prompts share the wrong answer, the less the shared policy suppresses it, and at $u=1$ the suppression disappears. Left: arrows from each condition's base model to its mean final hack fraction. Center: drift from the base model, zero under the filter baseline. Right: true accuracy. Dots are ten seeds, markers their means $\pm1$ s.e.m.\ at the operating point selected by verifier reward, and \emph{permuted blocks} randomizes block membership.}
\label{fig:universality}
\end{figure}

Our main results are demonstrated in Figure~\ref{fig:universality}, the \emph{headline sweep}: at a fixed false-positive rate, universality sets the direction for the shared-parameter policy. With the static false-positive rate fixed and only the number of prompts sharing the accepted wrong response varied, the shared-parameter policy's hack fraction climbs from $\NuTwoNeuralLoRho{}\pm\NuTwoNeuralLoRhoSem{}$ at $u=\nicefrac{1}{64}$ to $\NuTwoNeuralHiRho{}\pm\NuTwoNeuralHiRhoSem{}$ at $u=1$ (\NuTwoNSeeds{} seeds), against a base value of \NuTwoBaseRhoMean{}: as a drift, $\Delta\rho$ moves from $\NuTwoNeuralLoDR{}\pm\NuTwoNeuralLoDRSem{}$ to $\NuTwoNeuralHiDR{}\pm\NuTwoNeuralHiDRSem{}$, and true accuracy falls from \NuTwoNeuralLoAcc{} to \NuTwoNeuralHiAcc{}. The per-prompt policy, initialized identically and trained on the same grid, is flat: regressing $\Delta\rho$ on $\log_2 m$ gives $\NuTwoTabSlope{}$ per doubling against $\NuTwoNeuralSlope{}$ for the shared policy. It does not sit on the filter baseline, but its deviation carries no information about universality, as Corollary~\ref{cor:tabular} requires. Randomizing block membership reproduces both the suppression and the slope ($\NuSixSlope{}$ against $\NuSixTabSlope{}$ for the per-prompt arm), so contiguous blocks, which for $m\geq8$ group prompts by first operand $a$, are not the cause.

The $u=1$ end under Adam is heterogeneous: the mean drift at the selected operating point is within its own error bar, the ten final hack fractions span \NuTwoHiSeedList{}, with one collapsed seed and a median (\NuTwoHiSeedMedian{}) below the base value, and \NuTwoFlipHi{}\% of prompts finish with the leak more probable than the correct answer against \NuTwoFlipLo{}\% at $u=\nicefrac{1}{64}$, which is why $\Delta\rho$ approaches zero while $\Delta\Lambda$ stays negative ($\NuTwoNeuralLoDL{}$ to $\NuTwoNeuralHiDL{}$). A decisive sign change into amplification appears under sampled GRPO, under plain SGD (Section~\ref{sec:practice}), and under Adam at the next learning rate up (Section~\ref{sec:lr}). Crossing universality with the static false-positive rate, both swept eightfold at a fixed total base hack mass (the base model's total probability on leaks), attributes \NuThreeEtaM{} of the variance in $\Delta\Lambda$ to universality and \NuThreeEtaK{} to the false-positive rate (Figure~\ref{fig:field}, Appendix~\ref{app:extra}): the false-positive rate is the weaker variable.

\subsection{The slope holds at every convergent learning rate but the sign of the drift does not}
\label{sec:lr}

\begin{figure}[t]
\centering
\includegraphics[width=\linewidth]{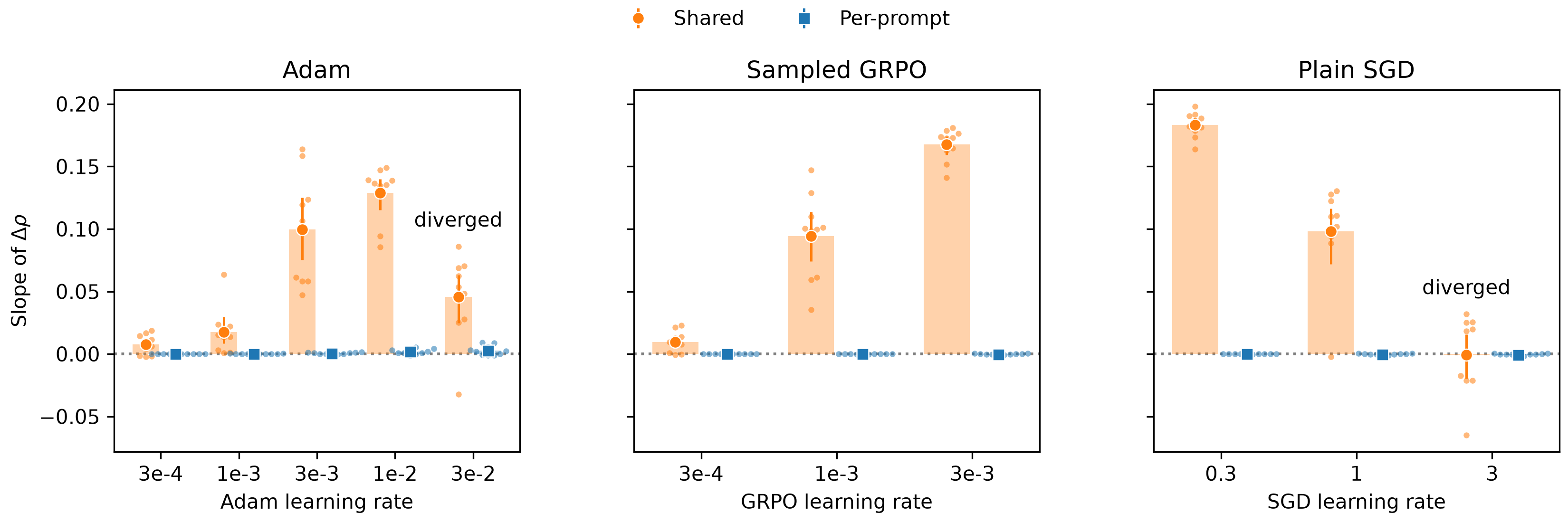}
\caption{The universality slope of $\Delta\rho$ (per doubling of $m$) at every learning rate, for both policies under three training setups. Bars and markers are means over ten seeds, dots are seeds, and whiskers bootstrap $95\%$ intervals. The shared slope is positive everywhere except where SGD diverges; the per-prompt slope is near zero everywhere.}
\label{fig:lr}
\end{figure}

Figure~\ref{fig:lr} reports the universality slope at every learning rate, for three training setups (Adam, sampled GRPO, and plain SGD) and both arms. The shared-parameter slope is positive with a bootstrap interval excluding zero at \LrNPosCI{} of \LrNTotal{} settings, the exception being the largest SGD learning rate, where training diverges. The per-prompt slope is within a few thousandths of zero at every setting: cross-prompt sharing adds a component to the drift that grows with universality at every convergent step size we tried.

The sign of the drift is not robust. Under Adam, all ten seeds suppress the idiosyncratic leak at the three smallest learning rates and all ten amplify it at the two largest, where accuracy on that condition also collapses (\LrAdamDAccLo{} at learning rate $10^{-2}$); the per-prompt arm's drift moves with step size in the same way, without a slope, so that movement is prompt-local. \LrAdamNSat{} Adam learning rates saturate the verifier at $u=1$ to within the operating-point rule's tolerance; the rule picks the smallest, learning rate $10^{-3}$, where $\Delta\rho$ at $u=1$ is $\LrAdamBDRHi{}$, whereas at learning rate $3\times10^{-3}$ it is $\LrAdamCDRHi{}\pm\LrAdamCDRHiSem{}$ with \LrAdamCNPosHi{} of \LrAdamCN{} seeds positive while the idiosyncratic leak is still suppressed in every seed ($\LrAdamCDRLo{}$, accuracy \LrAdamCAccLo{}). The suppression claim therefore carries a step-size qualifier, and the missing amplification at $u=1$ in Figure~\ref{fig:universality} reflects the conservative tie-break rather than the optimizer (Table~\ref{tab:lr}).

\subsection{Sharing is the channel}
\label{sec:mechanism}

\begin{figure}[t]
\centering
\includegraphics[width=\linewidth]{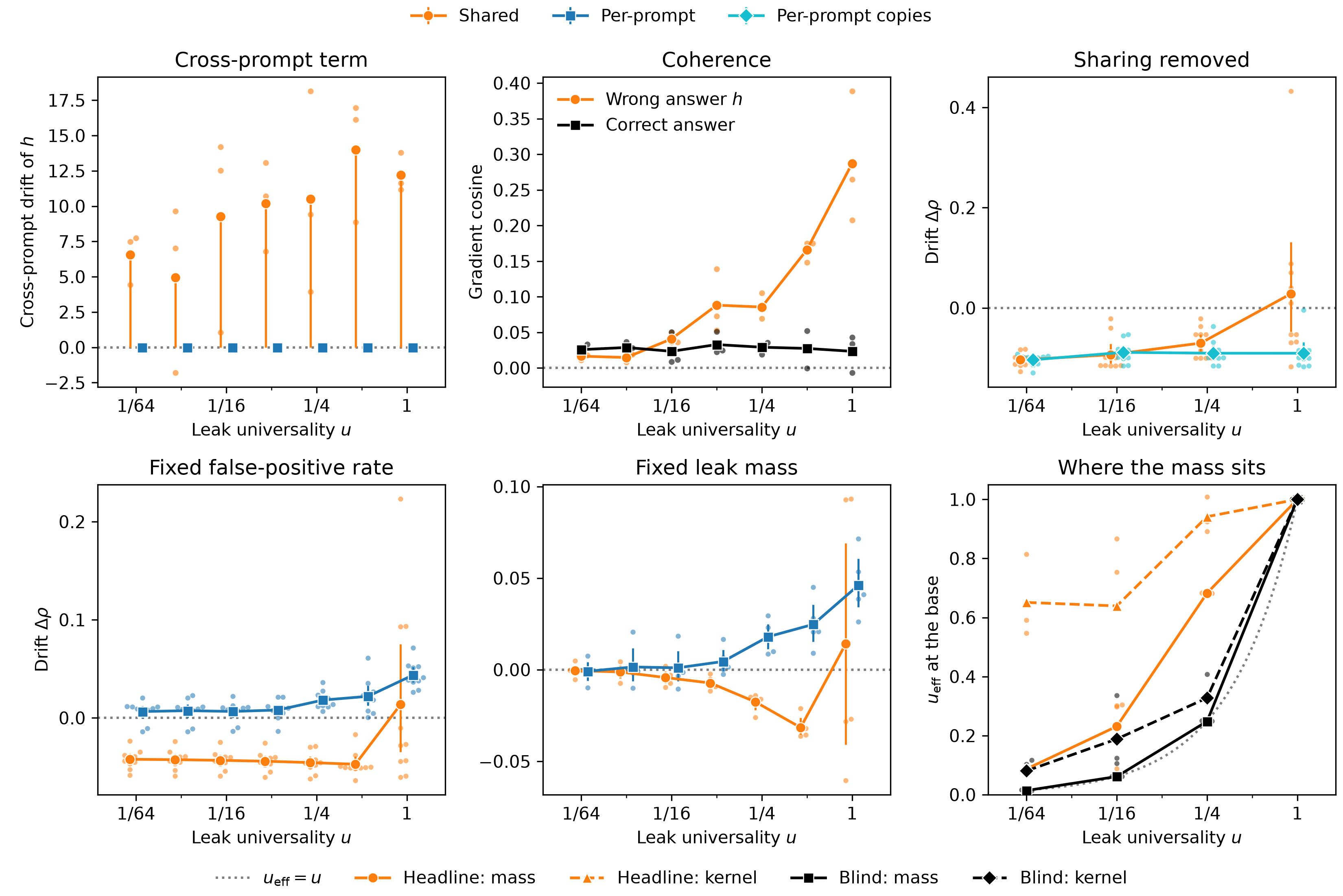}
\caption{Sharing is the channel (top), and the base model decides where the mass sits (bottom). Top: the cross-prompt component of the exact drift of $\log\pth(h\mid x)$ at the base model (Lemma~\ref{lem:drift}), zero for per-prompt parameters; the cross-prompt gradient cosine of each prompt's own leak and of its correct answer; and the shared transformer against per-prompt copies of the identical architecture. Bottom: one frozen base per seed with only the verifier's mask varying, holding either the static false-positive rate or the accepted string's per-prompt base mass fixed; and $\ueff$ (Equation~\ref{eq:ueff}) at the base model against raw $u$, in its mass-only and kernel-weighted readings, for the headline and the frozen blind bases. Dots are seeds; whiskers in the drift panels are bootstrap $95\%$ intervals.}
\label{fig:mechanism}
\end{figure}

Lemma~\ref{lem:drift} splits the drift into an own-prompt term and a cross-prompt term that is the entire contribution of sharing (Figure~\ref{fig:mechanism}, top). By autograd at the base model (three seeds) the cross-prompt term of the drift of $\log\pth(h\mid x)$ is \MechTabCross{} for the per-prompt policy at every universality, where Corollary~\ref{cor:tabular} is exact, and for the shared policy it rises from $\MechCrossLo{}\pm\MechCrossLoSem{}$ to $\MechCrossHi{}\pm\MechCrossHiSem{}$. The cross-prompt gradient cosine of each prompt's own leak (the mean cosine between the gradients $\nabla_\theta\log\pth(h_x\mid x)$ across prompts, with $h_x$ the leak accepted at $x$) rises from $\CohHackLo{}$ to $\CohHackHi{}$ across the sweep, against $\CohGold{}$ for correct answers.

The tabular comparison confounds parameterization with sharing. The cleanest test gives each prompt its own copy of the transformer, all copies starting from the same checkpoint and taking the same number of steps, with copy $x$ evaluating only prompt $x$. The dependence of $\Delta\rho$ on universality disappears: the slope falls to $\CopiesSlope{}$ per doubling (CI $\CopiesSlopeCI{}$) against $\SharedSlopeMatched{}$ for the shared policy under matched conditions (same $u$ values and seeds), a paired within-seed difference of $\CopiesPairedDiff{}$ (CI $\CopiesPairedDiffCI{}$, \CopiesPairedN{} seeds), and at $u=1$ the copies drift $\CopiesHiDR{}$ and keep \CopiesHiAcc{} accuracy where the shared policy loses both. This is Corollary~\ref{cor:tabular} with the architecture held fixed. The hack fraction cannot show what sharing adds at low $u$, where both arms sit at the floor set by the few prompts the leak wins, but the log-odds can: at $u=\nicefrac{1}{64}$ the paired difference in $\Delta\Lambda$ between the shared policy and the copies is $\CopiesPairedDLLo{}$ (CI $\CopiesPairedDLLoCI{}$), and at $u=1$ it is $\CopiesPairedDLHi{}$ (CI $\CopiesPairedDLHiCI{}$).

\subsection{The base model determines where the mass sits}
\label{sec:controls}

In the sweeps above, changing the verifier also changes the supervised target, so verifier universality and the universality of the base model's own error modes coincide by construction. To separate them we freeze one checkpoint per seed, the \emph{blind base}, whose supervised target carries a shared string $h$ and a per-prompt string $s_x$ on every prompt, identically in every condition, and vary only which prompts the verifier accepts $h$ on; one variant (\emph{fixed-rate}) accepts $s_x$ on the remaining prompts, which fixes the static false-positive rate, and the other (\emph{fixed-mass}) accepts nothing wrong there, which fixes the accepted string's per-prompt base mass (Appendix~\ref{app:details}).

Read per universality value, the shared policy's curve is a step, not a line (Figure~\ref{fig:mechanism}, bottom). In the fixed-rate variant the shared policy suppresses the accepted string far more than the per-prompt policy at every $u<1$, with a paired within-seed difference of $\MaskFprPairedLo{}$ at $u=\nicefrac{1}{64}$ (CI $\MaskFprPairedLoCI{}$) and $\MaskFprPairedMid{}$ at $u=\nicefrac12$ (CI $\MaskFprPairedMidCI{}$), excluding zero at all \MaskFprNBelowOne{} conditions below $u=1$. At $u=1$, where no prompt rejects $h$, the shared policy's mean drift is within its own error bar and the per-prompt policy amplifies it, and at the next learning rate up the shared policy amplifies the blind base's string about as strongly as the headline sweep's ($\MaskFprThreeHiDR{}$ against $\LrAdamCDRHi{}$ at the same learning rate). The fixed-mass variant agrees wherever the accepted string carries enough base mass to move ($\MaskMassPairedMid{}$ at $u=\nicefrac12$, CI $\MaskMassPairedMidCI{}$; $\MaskMassThreeHiDR{}$ at $u=1$ at the higher learning rate).

The two variants therefore agree, and their difference from the headline sweep is what Equation~\eqref{eq:threshold} predicts. In the blind base the model's mass on $h$ is spread nearly evenly over prompts, so the mass-only reading of $\ueff$ (Equation~\eqref{eq:ueff} with every kernel entry set to one) is close to $u$ (Figure~\ref{fig:mechanism}, bottom right): every rejecting prompt pushes $h$ down, and everything below $u=1$ ends suppressed once $\bar V$ has risen. In the headline sweep the supervised stage puts the leak's mass on the prompts that accept it and only the floor from the uniform component elsewhere, so the same reading is well above $u$ ($\UeffMassEOneLo{}$, $\UeffMassEOneFour{}$, $\UeffMassEOneSixteen{}$, and $\UeffMassEOneHi{}$ at $u = \nicefrac{1}{64}, \nicefrac{1}{16}, \nicefrac14$, and $1$; Equation~\eqref{eq:ueff} itself, the kernel-weighted reading, separates the two bases likewise), the push-down from rejecting prompts is weak, and the drift rises gradually with $u$. What the base model contributes is not a switch that enables sharing but the placement of its own mass; the two bases represent $h$ at $u=1$ with similar cross-prompt gradient cosine ($\RepENineCoh{}$ against $\RepEOneCoh{}$; Appendix~\ref{app:probe}).

\subsection{Robustness across GRPO, SGD, and KL penalties}
\label{sec:practice}

\begin{figure}[t]
\centering
\includegraphics[width=\linewidth]{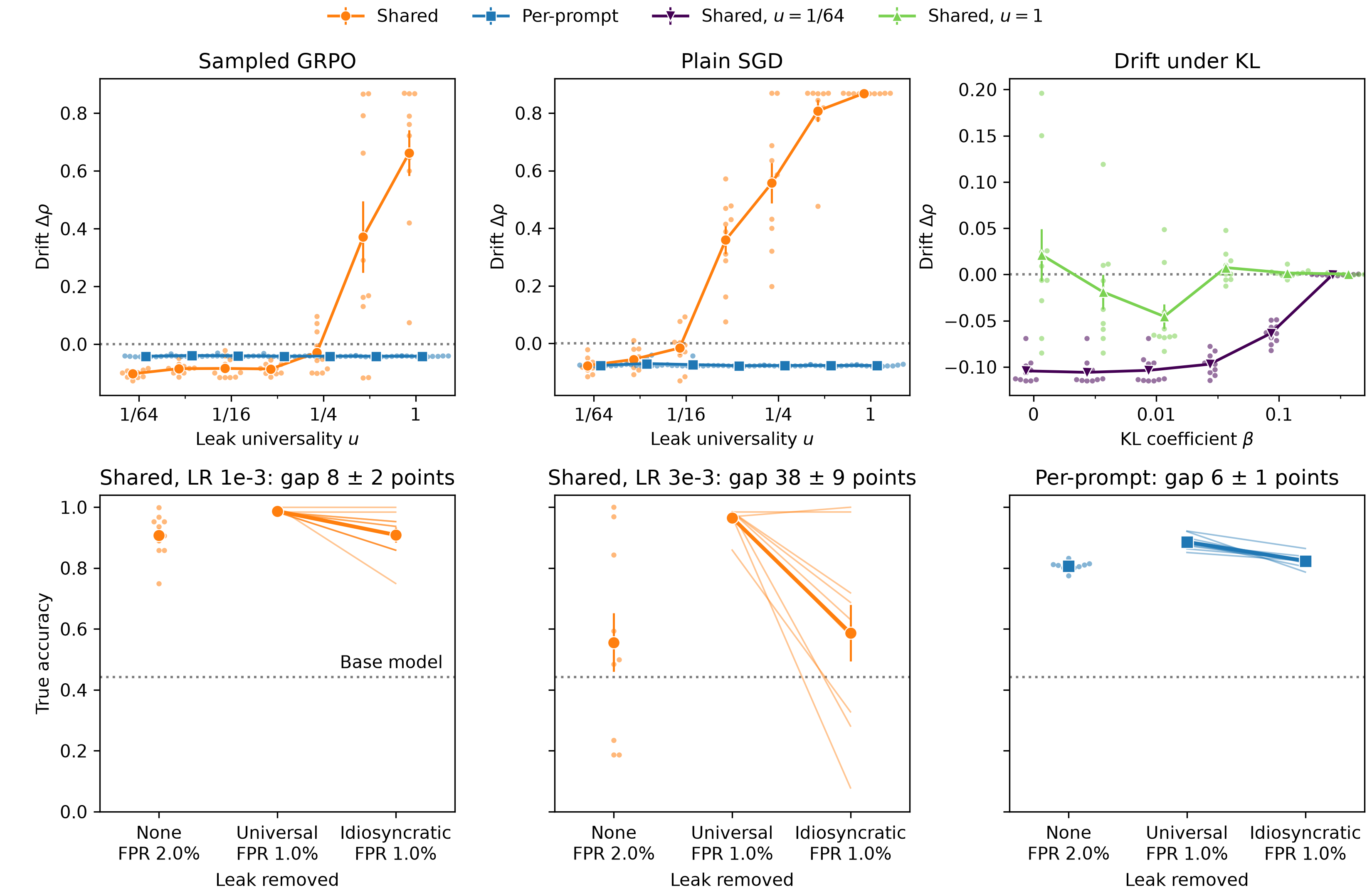}
\caption{The effect in practice. Top: sampled GRPO (group size \NGrpoG{}), plain SGD, and the drift at the two ends of the universality sweep as the KL coefficient $\beta$ grows (Adam, operating point); dots are seeds and markers their means $\pm1$ s.e.m. Bottom: two verifier patches with identical static audit value buy very different true accuracy; thin lines join one seed's two patched runs from the same checkpoint, titles give the paired gap with its s.e.m., and the dotted line is the base accuracy.}
\label{fig:practice}
\end{figure}

The effect is not an artifact of exact-expectation gradients or of Adam (Figure~\ref{fig:practice}, top). Under sampled GRPO \citep{shao2024deepseekmath} with group size \NGrpoG{} and standard-deviation normalization, $\Delta\rho$ moves from $\NuFiveADLo{}\pm\NuFiveADLoSem{}$ to $\NuFiveADHi{}\pm\NuFiveADHiSem{}$ over \NuFiveANSeeds{} seeds, crossing zero at \NuFiveACrossU{}. Under plain SGD, the optimizer the analysis assumes, the shared policy's hack fraction climbs from \NuTwoSgdNeuralLoRho{} to \NuTwoSgdNeuralHiRho{} and its accuracy falls from \NuTwoSgdNeuralLoAcc{} to \NuTwoSgdNeuralHiAcc{}, crossing at \NuTwoSgdCrossU{}, while the per-prompt policy stays below its base model throughout.

A KL penalty pulls the policy back toward the filter baseline, as Proposition~\ref{prop:filter} requires, and here that costs accuracy: at $u=\nicefrac{1}{64}$, raising $\beta$ to \NuFiveBBeta{} takes $\Delta\rho$ from $\NuFiveBDRNoKLLo{}$ to $\NuFiveBDRKLLo{}$, removing the beneficial suppression, and accuracy from \NuFiveBAccNoKLLo{} to \NuFiveBAccKLLo{} (\NuFiveBNSeeds{} seeds; Figure~\ref{fig:appendix}). KL is not a free defense: it commits the model to the base model's composition of the accepted set, which helps only when that composition is better than what training would reach; \citet{ackermann2026gradreg} reach a compatible conclusion from the other direction.

\subsection{Budgeting verifier hardening}
\label{sec:hardening}

We contaminate a base model with one universal and one idiosyncratic leak string per prompt, and compare two verifier patches, one removing the universal string and one every idiosyncratic string, that remove the same number of accepted wrong pairs of prompts and responses, so that a static audit rates them equally (Figure~\ref{fig:practice}, bottom). Both patches start from the same checkpoint within each seed, so every difference arises during RL, and the verifier's own reward stays at \NuFiveCLkyMin{} or above under both patches at both learning rates, so neither would be flagged during training. At the operating point (learning rate $10^{-3}$), the universal patch leaves the shared policy at \NuFiveCAccU{} true accuracy and the idiosyncratic patch at \NuFiveCAccI{}, a paired within-seed gap of $\NuFiveCGap{}\pm\NuFiveCGapSem{}$ points (bootstrap CI $\NuFiveCGapCI{}$, \NuFiveCNSeeds{} seeds); at learning rate $3\times10^{-3}$, where the universal leak is amplified, the gap is $\NuFiveCGapHiLr{}\pm\NuFiveCGapHiLrSem{}$ points (CI $\NuFiveCGapHiLrCI{}$, positive in \NuFiveCGapHiLrNPos{} of \NuFiveCNSeeds{} seeds). The per-prompt policy shows a gap of $\NuFiveCGapTab{}$ points (CI $\NuFiveCGapTabCI{}$), which can only come from prompt-local dynamics acting on the base model's unequal mass on the two kinds of leak; if the two effects add, the remaining $\NuFiveCGapRLHiLr{}$ points at the larger learning rate (CI $\NuFiveCGapRLHiLrCI{}$) require sharing, whereas at learning rate $10^{-3}$ the remainder is $\NuFiveCGapRL{}$ points (CI $\NuFiveCGapRLCI{}$). A patch that removes the error the model shares across prompts can be worth far more than one that removes the same number of scattered errors, and a static audit cannot tell the two apart.

\section{Discussion}

\paragraph{Conclusion.} Whether RLVR inherits or amplifies a base model's reward hacking has no single answer. RLVR is a neutral filter at its KL-regularized optimum, and cross-prompt parameter sharing breaks that baseline in both directions, through a channel proportional to the model's own mass on a wrong answer at the prompts that accept and reject it; removing the sharing removes the channel. In our environment, suppression and amplification are the two ends of one axis with inheritance between them, and at a given step size what moves a setting along it is how much of that mass sits on prompts where the verifier accepts it. For the two literatures, our results suggest a hypothesis rather than establish a reconciliation: the prior studies plausibly sit at different points, since the false positives in the leaky test suites of \citet{zhang2026leaky} are per-task while the hack in \citet{khalifa2026countdown} edits the test file and works everywhere, but no estimate of $u$ or $\ueff$ exists for those settings. \citet{zhang2026leaky} finds no per-task growth in rewarded false positives, which is inheritance rather than the suppression at our low-$u$ end, so its released rollouts are a natural place to estimate $\ueff$. What to audit is not just how many errors a verifier makes or how many prompts share them, but whether the model is wrong the same way the verifier is; for any accepted string, $\bar V$ and $\ueff$ depend only on the policy and the verifier's mask.

\paragraph{Limitations.} Proposition~\ref{prop:threshold} covers only one channel of the drift and assumes the same pass rate on every prompt. Other channels are about as large (Appendix~\ref{app:probe}), so it indicates a direction rather than giving a bound. The setting is small: exact enumeration limits responses to two tokens, so a shared error is one identical string rather than a shared way of going wrong, the leaks are injected through contaminated supervised training rather than discovered by the policy, and the policy is one two-layer transformer with \NParams{} parameters. The sign of the drift at low universality depends on the step size, and at $u=1$ under Adam it varies across seeds. Our reading of prior language model results is a hypothesis that we have not tested on those models.

\paragraph{Future work.} The natural next step is to scale up experiments. This would entail measuring $\ueff$ and the cross-prompt coherence of accepted wrong responses on a real language model and verifier, both of which can be computed from per-prompt gradients, and to test whether they predict which false positives RLVR amplifies.

\paragraph{AI usage.} LLM agents assisted with experiment design and manuscript preparation, but all code and text have been reviewed by the authors.

\RemoveFromHook{para/begin}[loose]

\bibliographystyle{plainnat}
\bibliography{references}

\appendix
\section{Proofs}
\label{app:proofs}

\subsection{Proposition~\ref{prop:filter}}

Fix $x$ and assume $\pi_{\mathrm{ref}}(\cdot\mid x)>0$, so the KL term is finite and every policy on the path is strictly positive. The objective $\sum_y \pi(y)r(x,y) - \beta\sum_y \pi(y)\log\frac{\pi(y)}{\pi_{\mathrm{ref}}(y)}$ over the simplex is strictly concave for $\beta>0$; setting the derivative of the Lagrangian to zero gives $\pi^\star_\beta(y\mid x) = \pi_{\mathrm{ref}}(y\mid x)e^{r(x,y)/\beta}/Z_\beta(x)$ with $Z_\beta(x)=\sum_{y'}\pi_{\mathrm{ref}}(y'\mid x)e^{r(x,y')/\beta}$. For $y,y'$ with $r(x,y)=r(x,y')$ the normalizer and the exponential factor cancel in the ratio. Since $r\geq c$ we have $\Hset(x)\subseteq\Acc(x)$ and $r\equiv 1$ on $\Acc(x)$, so
\[
\frac{\pi^\star_\beta(\Hset(x)\mid x)}{\pi^\star_\beta(\Acc(x)\mid x)}
= \frac{e^{1/\beta}\pi_{\mathrm{ref}}(\Hset(x)\mid x)/Z_\beta(x)}{e^{1/\beta}\pi_{\mathrm{ref}}(\Acc(x)\mid x)/Z_\beta(x)}
= \frac{\pi_{\mathrm{ref}}(\Hset(x)\mid x)}{\pi_{\mathrm{ref}}(\Acc(x)\mid x)}
\]
for every $x$, hence $\rho$ and $\Lambda$ are unchanged. For a multiplicative update $\pi_{t+1}(y\mid x)\propto\pi_t(y\mid x)e^{\eta f(r(x,y))}$ the same cancellation applies at every step.

\paragraph{Excursion of the pooled statistic.} Write $a_x = \pi_{\mathrm{ref}}(\Acc(x)\mid x)$, $\rho_x = \pi_{\mathrm{ref}}(\Hset(x)\mid x)/a_x$, and $s=e^{1/\beta}$. Then $\pi^\star_\beta(\Acc(x)\mid x) = s\,a_x/(1-a_x+s\,a_x)$, and the pooled statistic is $P(s) = \sum_x \rho_x w_x(s)/\sum_x w_x(s)$ with $w_x(s)=s a_x/(1-a_x+s a_x)$. At $s=1$ (no RL) $P = \sum_x a_x\rho_x/\sum_x a_x$; as $s\to\infty$ every $w_x\to1$ provided $a_x>0$, and $P \to \frac1N\sum_x \rho_x = \rho(\pi_{\mathrm{ref}})$. The total excursion is therefore
\[
P(\infty)-P(1) \;=\; \frac1N\sum_x \rho_x \;-\; \frac{\sum_x a_x\rho_x}{\sum_x a_x}
\;=\; -\,\frac{\Cov_x(a_x,\rho_x)}{\mathbb{E}_x[a_x]},
\]
which is positive whenever pass rate and hack fraction are negatively correlated across prompts. It vanishes when $\Cov_x(a_x,\rho_x)=0$; constant $\rho_x$ is sufficient but not necessary, since equal $a_x$ at every prompt also gives $P(s)=\frac1N\sum_x\rho_x$ for every $s$. The pooled statistic is not an estimator of $\rho$; it is a deterministic function of the filtering path that happens to converge to $\rho$ in the strong-RL limit.

\subsection{Lemma~\ref{lem:drift}}

$\nabla_\theta J = \frac1N\sum_{x'}\sum_{y'}\pth(y'\mid x')r(x',y')\nabla_\theta\log\pth(y'\mid x') = \frac1N\sum_{x'}\mathbb{E}_{y'}[r(x',y')\nabla_\theta\log\pth(y'\mid x')]$. Since $\mathbb{E}_{y'}[\nabla_\theta\log\pth(y'\mid x')]=0$, the expectation equals $\Cov_{y'}(r,\nabla_\theta\log\pth)$ componentwise. For a fixed vector $v=\nabla_\theta\log\pth(y\mid x)$, $\langle v,\Cov(r,g)\rangle = \mathbb{E}[r\langle v,g\rangle]-\mathbb{E}[r]\mathbb{E}[\langle v,g\rangle]=\Cov(r,\langle v,g\rangle)$, so the inner product passes inside. \qed

\subsection{Corollary~\ref{cor:tabular}}

With $\theta=(\theta_x)$ and $\pth(\cdot\mid x)$ a function of $\theta_x$ only, $\nabla_{\theta_{x'}}\log\pth(y\mid x)=0$ for $x'\neq x$, so every off-diagonal kernel entry is a sum of products with a zero factor. Moreover $\nabla_{\theta_x}J=\frac1N\sum_y\pth(y\mid x)(r(x,y)-V(x))\nabla_{\theta_x}\log\pth(y\mid x)$ involves only $\theta_x$, so the flow restricted to block $x$ is autonomous and depends on the verifier only through $r(x,\cdot)$. The argument survives Adam, whose state is per-coordinate and whose momentum is a linear combination of gradients that are already block-supported; it would not survive a global gradient-norm clip, which we do not use. \qed

\paragraph{Rich-get-richer under vanilla policy gradient.} For a single prompt with free logits $\theta_y$ and $J=\sum_y\pi_y r_y$, $\partial J/\partial\theta_z = \pi_z(r_z-J)$, so for two responses with $r_1=r_2=1$, $\frac{d}{dt}\log(\pi_1/\pi_2)=(\pi_1-\pi_2)(1-J)$. In the $N$-prompt setting of Corollary~\ref{cor:tabular} the objective carries a $1/N$ and the baseline is the prompt's own pass rate, giving $\frac1N(\pi_1-\pi_2)(1-V(x))$. The odds between equally rewarded responses are therefore not preserved by vanilla policy gradient even without sharing; they move toward whichever response already has more mass, at a rate that vanishes only as $V(x)\to1$. This drift is independent of $u$ by Corollary~\ref{cor:tabular}. It is a statement about gradient flow: under Adam a leak that starts behind the correct answer can overtake it, which this drift cannot do, so the per-prompt arm's movement with step size in Section~\ref{sec:lr} is prompt-local but is not this effect.

\subsection{Proposition~\ref{prop:threshold}}

Two parts of the kernel drop out before Definition~\ref{def:coherent} is applied: a component of $\Kk((x,h),(x',y'))$ that does not depend on $y'$ contributes zero covariance in \eqref{eq:drift}, and a component that does not depend on $h$ is common to the leak and to the correct answer at prompt $x$ and cancels in the within-prompt log-odds. Substituting the same-string part of the definition, the covariance at $x'$ is
\[
\Cov_{y'}[r(x',y'),\kappa_{xx'}\mathbf{1}[y'=h]] = \kappa_{xx'}\,\pth(h\mid x')\big(r(x',h)-V(x')\big),
\]
using $\mathbb{E}_{y'}[r]=V(x')$. Under $V(\cdot)\equiv\bar V$ and with $r(x',h)=\mathbf{1}[x'\in U]$,
\[
\begin{aligned}
\frac{d}{dt}\log\pth(h\mid x)\Big|_{\mathrm{coh}}
&=\frac{1}{N}\Big[\sum_{x'\in U}\kappa_{xx'}\pth(h\mid x')(1-\bar V)-\sum_{x'\notin U}\kappa_{xx'}\pth(h\mid x')\bar V\Big]\\
&=\frac{W_x}{N}\big(\ueff(x,h)-\bar V\big),
\end{aligned}
\]
with $W_x=\sum_{x'}\kappa_{xx'}\pth(h\mid x')$ and $\ueff$ as in \eqref{eq:ueff} with $\kappa_{xx'}$ in place of $\Kxx$; the sum includes $x'=x$, which is the own-prompt term. The identical computation for the correct answer $g_x$ with $\kappa^g_{xx'}=0$ off the diagonal leaves only its own term, $\frac1N\kappa_{s,g}\pth(g_x\mid x)(1-\bar V)$, because $r(x,g_x)=1$. Subtracting gives \eqref{eq:threshold}. Under the further assumptions $x\in U$, $\pth(h\mid\cdot)\equiv p_h$, and $\kappa_{xx'}=\kappa$ off the diagonal, $\kappa_s$ on it, $W_x=p_h[\kappa_s+(N-1)\kappa]$ and $\ueff=[\kappa_s+(m-1)\kappa]/[\kappa_s+(N-1)\kappa]$, so
\[
\frac{W_x}{N}(\ueff-\bar V)=\frac{p_h}{N}\big[\kappa(m-1)-\kappa\bar V(N-1)+\kappa_s(1-\bar V)\big]
=\kappa p_h(u-\bar V)+\frac{(1-\bar V)}{N}(\kappa_s-\kappa)p_h,
\]
which with the correct answer's own term is the homogeneous form quoted in the proposition. For $x\notin U$ the same computation gives the same leading term with $(1-\bar V)$ replaced by $-\bar V$ in the correction term. \qed

\paragraph{What the proposition does and does not say.} The correction in the homogeneous form is $O(1/N)$ at fixed kernel scales, but it is small relative to the leading term when $(1-\bar V)\big(|\kappa_s-\kappa|p_h+\kappa_{s,g}p_g\big)\ll N\kappa p_h|u-\bar V|$, which constrains the kernel, the correct answer's own self-scale, and the distance from the threshold, not $N$ alone; this is sufficient rather than necessary, since the two parts of the correction can cancel, and it says nothing at $u\approx\bar V$. The binding factor in our environment is $N\kappa/\kappa_s$, with $\kappa/\kappa_s$ measured as the mean cross-prompt gradient cosine of each prompt's own leak: \CohHackHiFine{} at $u=1$ gives $N\kappa/\kappa_s\approx\NThreshValidity{}$, but \CohHackLoFine{} at $u=\nicefrac{1}{64}$ gives about $\NThreshValidityLo{}$, where the correction is the same order as the leading term (and at low $u$ the leaks are different strings, so this is not even the same-string scale that the proposition uses). The general form \eqref{eq:threshold} does not need homogeneity of the masses, which is the form we use in Section~\ref{sec:controls}, but it still describes only the same-string channel: the residual $R$ in Definition~\ref{def:coherent}, which includes the cross-prompt effect of the correct answers' shared machinery on the leak's logit, is not bounded. Appendix~\ref{app:probe} measures how much of the exact cross-prompt drift the channel accounts for.

\paragraph{Reading the threshold.} $\bar V$ is the policy's verifier pass rate, observable during training without ground truth, and $\ueff$ depends only on the policy and the verifier's mask. A coherent wrong answer is pushed up by the same-string channel when the share of the policy's kernel-weighted mass on it that sits on accepting prompts exceeds the policy's current pass rate. The statement is directional and first-order; it identifies the sign of one channel and the quantity controlling it, not an exact boundary in a model whose kernel is only approximately coherent.

\section{Environment, Model, and Training}
\label{app:details}

\paragraph{Task.} Prompts are all $(a,b)$ with $a,b\in\{2,\dots,9\}$ ($N=64$); the correct response is the two-digit decimal string of $ab$ (all products lie in $[4,81]$). The response space is all $100$ two-digit strings. Reducing the operand set to $\{2,\dots,5\}$ gives $N=16$ for the prompt-set-size sweep. The correct answers are not perfectly prompt-specific ($2\times6$ and $3\times4$ both give $12$), so they carry a universality of their own, $u_g=\NGoldU{}$. Under Definition~\ref{def:coherent} what matters is measured coherence rather than string identity, and we find the correct responses to be near-incoherent across prompts (\CohGold{}) while a universal leak is not.

\paragraph{Leak assignment.} Leak strings are drawn without replacement from the $69$ two-digit strings that are not the correct answer of any prompt, so a leak is never a correct answer anywhere and every response is unambiguously classified. Prompts are partitioned into $N/m$ blocks; each block receives $k$ leak strings shared by all its prompts. The verifier's static false-positive rate is exactly $k/99$ in every condition, independent of $m$. Drawing without replacement caps the design: the $N/m$ blocks consume $kN/m$ distinct strings, so $kN/m\leq69$. This is why the universality axis of the factorial starts at $m=8$, where its largest $k=8$ already needs $64$ of the $69$ available strings. The single-$k$ sweeps are unconstrained.

Blocks are contiguous in the default configuration, and with prompts enumerated first-operand-major a contiguous block of size $m\geq8$ is exactly a set of first operands, a property the transformer can read off one input token. That would confound universality with how simple the prompt-to-leak map is to represent, so we run the whole sweep a second time with block membership given by a random permutation of prompts and leak strings drawn independently per $(m,k,\text{seed})$. Both versions appear in Figure~\ref{fig:universality}.

\paragraph{What the base model does and does not control.} The supervised target puts the same hack mass on a given prompt in every condition, but the realized base models differ slightly, so the measured base hack fraction still varies by \NuTwoBaseRhoSpread{} across universality conditions. We therefore report drifts from each run's own base model, which Proposition~\ref{prop:filter} makes exactly zero under the filter baseline. In the universality$\times$false-positive-rate factorial the uniform component of the supervised target is spread only over responses that are neither the correct answer nor a leak, which makes total base hack mass exactly $\varepsilon$ for every $k$; otherwise base hack mass would grow with $k$ and the two factors would not be separable. The measured base \emph{fraction} still varies across runs, mean $\rho = \NuThreeBaseRhoLoK{}$ with spread \NuThreeBaseRhoSpreadK{}. This is the only experiment with a different uniform component, and it is also why the mild decrease of the drift with $k$ in Figure~\ref{fig:field} is a property of the design: the same base hack mass is divided among more accepted strings.

\paragraph{Base model.} A decoder-only transformer with $2$ layers, model dimension $64$, $4$ heads, and a $10$-way output head over digits (\NParams{} parameters) is trained for $600$ Adam steps at $3\times10^{-3}$ on the exact target $q(y\mid x)=(1-\varepsilon)[\lambda_x\mathbf{1}[y=g_x]+(1-\lambda_x)\mathrm{Unif}(\Yset)]+\varepsilon\,\mathrm{Unif}(\Hset(x))$ with $\varepsilon=0.03$. The values $\lambda_x$ are an evenly spaced grid on $[0,0.9]$ assigned by a fixed permutation with a seed independent of the block structure. Because the loss is against an exact target rather than samples, the target is matched across universality conditions without per-condition calibration. Note that the uniform component puts a floor of $(1-\varepsilon)(1-\lambda_x)/100$ on every string at every prompt, including a block's leak string on prompts outside the block; that floor is what makes $\ueff$ smaller than one for $u<1$ in the headline sweep (Appendix~\ref{app:probe}).

\paragraph{Per-prompt policy.} The tabular policy has free parameters $w_1\in\mathbb{R}^{N\times10}$ and $w_2\in\mathbb{R}^{N\times10\times10}$ with $\pi(y\mid x)=\mathrm{softmax}(w_1[x])_{y_1}\cdot\mathrm{softmax}(w_2[x,y_1])_{y_2}$. The factorization is complete: its $9+10\cdot9=99$ free directions match the dimension of the simplex over $100$ responses. Given the base transformer's exact log-probabilities we set $w_1 = \log\sum_{y_2}\pi(y\mid x)$ and $w_2 = \log\pi(y\mid x) - w_1$, which reproduces the base distribution to within $10^{-4}$ in max-norm (asserted at run time).

\paragraph{RL.} The default objective is the exact-expectation policy gradient with the exact mean baseline, $\nabla_\theta J = \frac1N\sum_x\sum_y\pth(y\mid x)\,[r(x,y)-V(x)]\,\nabla_\theta\log\pth(y\mid x)$, implemented as a weighted cross-entropy with detached weights. The sampled-GRPO experiment uses group size $8$, group-standard-deviation normalization, and one inner epoch. Training uses Adam (default) or SGD for $400$ steps, with the learning-rate grid $\{3\times10^{-4},10^{-3},3\times10^{-3},10^{-2},3\times10^{-2}\}$ for Adam and $\{0.3,1,3\}$ for SGD, applied identically to both arms where a grid is shared (Table~\ref{tab:protocol}). The reported operating point is, for each arm, cell, and seed, the smallest learning rate whose final verifier reward is within $10^{-3}$ of the best. We apply this one rule to every experiment, including verifier hardening.

\paragraph{The mask-only controls (E9, E9b).} One checkpoint per seed, the blind base of Section~\ref{sec:controls}, serves every universality condition. Its supervised target is contaminated with a shared string $h$ and a per-prompt string $s_x$ at every prompt, in identical proportion in all conditions, so the initial policy, the probability on each leak string, and the initial tangent kernel are bit-identical across the sweep. The verifier then accepts $h$ on the $m$ prompts in $U$ and, in E9 (the fixed-rate variant), $s_x$ on the rest, giving exactly $N$ accepted wrong pairs and a static false-positive rate of exactly $1/(|\Yset|-1)$ in every condition. E9b (the fixed-mass variant) accepts nothing wrong outside $U$, so every accepted wrong string is $h$ and its per-prompt base mass is identical across conditions, at the cost of a static false-positive rate of $u/99$ and a base hack fraction that scales with $u$ (prompts that accept nothing wrong have $\rho_x=0$). $U$ is nested across $m$. The residual gradient in E9 is that the accepted string's base mass rises from \MaskFprBaseLo{} to \MaskFprBaseHi{} with $u$, because a shared-parameter network represents one string demonstrated at every prompt better than $N$ distinct ones; the per-prompt arm responds to that gradient through prompt-local dynamics, which Corollary~\ref{cor:tabular} permits since it is inherited rather than generated. Reweighting the two families in the supervised target until their realized probabilities match closes the mean gap but not the per-prompt gap, because the shared network cannot represent an arbitrary per-prompt split; we therefore bracket the confound with two designs rather than try to eliminate it.

\paragraph{Per-prompt weight copies (E10).} Each of $N$ independent copies of the same transformer is initialized from the shared base checkpoint and trained on its own prompt alone with the same objective and optimizer, at the three smallest learning rates of the shared arm's grid. Copy $x$ only ever evaluates prompt $x$, so training the copies in one loop is equivalent to training them separately. The matched shared-parameter arm is E1 restricted to the same four universality conditions and the same ten seeds, with slopes averaged per seed, which is why its slope (\SharedSlopeMatched{}) differs slightly from the seven-condition figure (\NuTwoNeuralSlope{}).

\paragraph{Verifier hardening (E6).} The base model is contaminated with both leak families, one universal string and one idiosyncratic string per prompt. Three verifiers are then deployed from the same checkpoint: unpatched, universal leak removed, and all idiosyncratic leaks removed. The two patches remove exactly $N$ accepted (prompt, wrong-response) pairs each. Both arms are trained at two learning rates ($10^{-3}$ and $3\times10^{-3}$ for the shared arm, thirty times that for the per-prompt arm, as in the KL sweep), and the operating point is chosen by the paper's standard rule, which picks $10^{-3}$ for the shared arm. The gaps in Section~\ref{sec:hardening} are paired within-seed differences between the two patched verifiers over \NuFiveCNSeeds{} seeds.

\paragraph{The threshold probe (E11).} For each probed prompt $x$ that accepts a leak $h$ we compute by autograd the exact drift of $\log\pth(h\mid x)-\log\pth(g_x\mid x)$ under gradient flow, split into its own-prompt and cross-prompt parts, and the gradients of $\log\pth(h\mid x')$ and $\log\pth(g_x\mid x')$ at every prompt $x'$, which give the same-string kernel rows $\Kxx$ and the probe's estimate of $\ueff$ (Equation~\eqref{eq:ueff}, with the own term included). The coherent-channel prediction is the $y'=h$ and $y'=g_x$ terms of Lemma~\ref{lem:drift} summed over $x'\neq x$, with the measured masses and kernel entries; the homogeneous formula uses the mean off-diagonal entry as $\kappa$, the diagonal entry as $\kappa_s$, and the leak's mean probability on accepting prompts as $p_h$. The probe runs at the base and at SGD steps \ProbeSteps{} with step size $0.3$, on \ProbeCap{} accepting prompts per condition at $m\in\{1,4,16,64\}$ and \ProbeNSeeds{} seeds, and once at the frozen blind base of E9b for the same $m$.

\paragraph{Full specification.} Prompts are serialized as the token sequence $[\textsc{bos}, a, b, y_1]$ over a vocabulary of $11$ symbols, with learned token and positional embeddings, a causal mask, pre-norm blocks, $4$ heads, feed-forward width $256$ with GELU, final \texttt{LayerNorm}, and a $10$-way output head shared across the two response positions. Embeddings are initialized $\mathcal{N}(0, 0.02^2)$ and the remaining parameters use PyTorch defaults. Optimization is full-batch over all $N$ prompts with no minibatching and no gradient clipping. Adam uses $\beta_1=0.9$, $\beta_2=0.999$, $\varepsilon=10^{-8}$, and zero weight decay; SGD uses no momentum. A run's seed sets the transformer initialization, the leak-string draw, and, in E4, the rollout sampling; the permutation of the $\lambda_x$ uses a separate fixed seed. E4 draws $G=8$ responses per prompt at temperature $1$ from the enumerated distribution, normalizes advantages by the group standard deviation (plus $10^{-4}$), and takes one inner epoch with no importance ratio, clipping, or reference-KL term. E5 adds $-\beta\,\mathrm{KL}(\pth(\cdot\mid x)\Vert\pi_{\mathrm{ref}}(\cdot\mid x))$ computed exactly against the frozen base.

\paragraph{Measurement.} $\rho$, $\Lambda$, and true accuracy are computed from the exact enumerated distribution over all $100$ responses at all $N$ prompts at every logged step; the only stochasticity is over seeds. The gradient decomposition and coherence probes use three seeds and average over fixed random subsamples of prompts, for cost. Per-prompt log-odds are clipped to $\pm20$ before averaging, so a prompt whose leak mass has collapsed cannot dominate $\Lambda$; at the operating point of the headline sweep this binds on a median of $1.6\%$ of prompts for the shared policy, and the ordering of every reported series is unchanged at clipping thresholds of $20$, $40$, $60$, and $200$. The final metric of every run is recomputed in float64 from stored per-prompt masses. The two null-model checks of Section~\ref{sec:design} are floating-point zeros: exact mirror descent preserves $\rho$ to \NullDriftMirror{} and $\Lambda$ to \NullDriftMirrorLam{} over \NullMirrorSteps{} steps, and the KL-regularized optimum preserves $\rho$ to \NullDriftKL{} across $\beta \in \NullBetaRange{}$. The two agree to the last bit by construction, since $400$ steps of mirror descent at $\eta=0.05$ reach exactly $\pi_0 e^{20 r}$, the $\beta=0.05$ endpoint of the KL grid. The per-prompt policy's cross-prompt kernel is \KernelRatioTab{}, against \KernelRatioNeural{} of the self-kernel for the shared policy.

\begin{table}[h]
\centering\footnotesize
\setlength{\tabcolsep}{4pt}\begin{tabular}{llcc@{\hspace{4pt}}l}
\toprule
 & sweep & seeds & learning rates & notes \\
\midrule
E1 & universality, contiguous blocks & 10 & 5, shared by both arms & headline sweep \\
E2 & universality, permuted blocks   & 10 & 3, shared by both arms & base spread \NuSixBaseSpread{} \\
E3 & universality $\times$ false-positive rate & 10 & 1 pair ($10^{-3}$ / $3\!\times\!10^{-2}$) & clean uniform floor \\
E4 & sampled GRPO                     & 10 & 3, shared by both arms & $G=\NGrpoG{}$, std norm. \\
E5 & KL sweep                         & 10 & 1 pair ($10^{-3}$ / $3\!\times\!10^{-2}$) & 6 values of $\beta$ \\
E6 & verifier hardening               & 10 & 2 pairs & standard rule \\
E7 & prompt-set size $N\in\{16,64\}$   & 10 & 3, shared by both arms & tests $u$ versus $m$ \\
E8 & plain SGD                        & 10 & 3, shared by both arms & robustness \\
E9 & mask-only, fixed false-positive rate & 10 & 5, shared by both arms & one frozen base per seed \\
E9b & mask-only, fixed leak mass & 5 & 5, shared by both arms & one frozen base per seed \\
E10 & per-prompt weight copies         & 10 & 3 & same architecture \\
E11 & threshold probe                  & 3 & SGD $0.3$ & exact vs.\ coherent drift \\
\bottomrule
\end{tabular}
\caption{What was run in each experiment. Where a grid is shared by both arms the operating point is the smallest learning rate within $10^{-3}$ of the best verifier reward, chosen independently for each arm. E2, E3, E5, E6, E7, and E11 were run from base models regenerated with the released code (Appendix~\ref{app:repro}).}
\label{tab:protocol}
\end{table}

\section{The Threshold Probe}
\label{app:probe}

\begin{figure}[htbp]
\centering
\includegraphics[width=\linewidth]{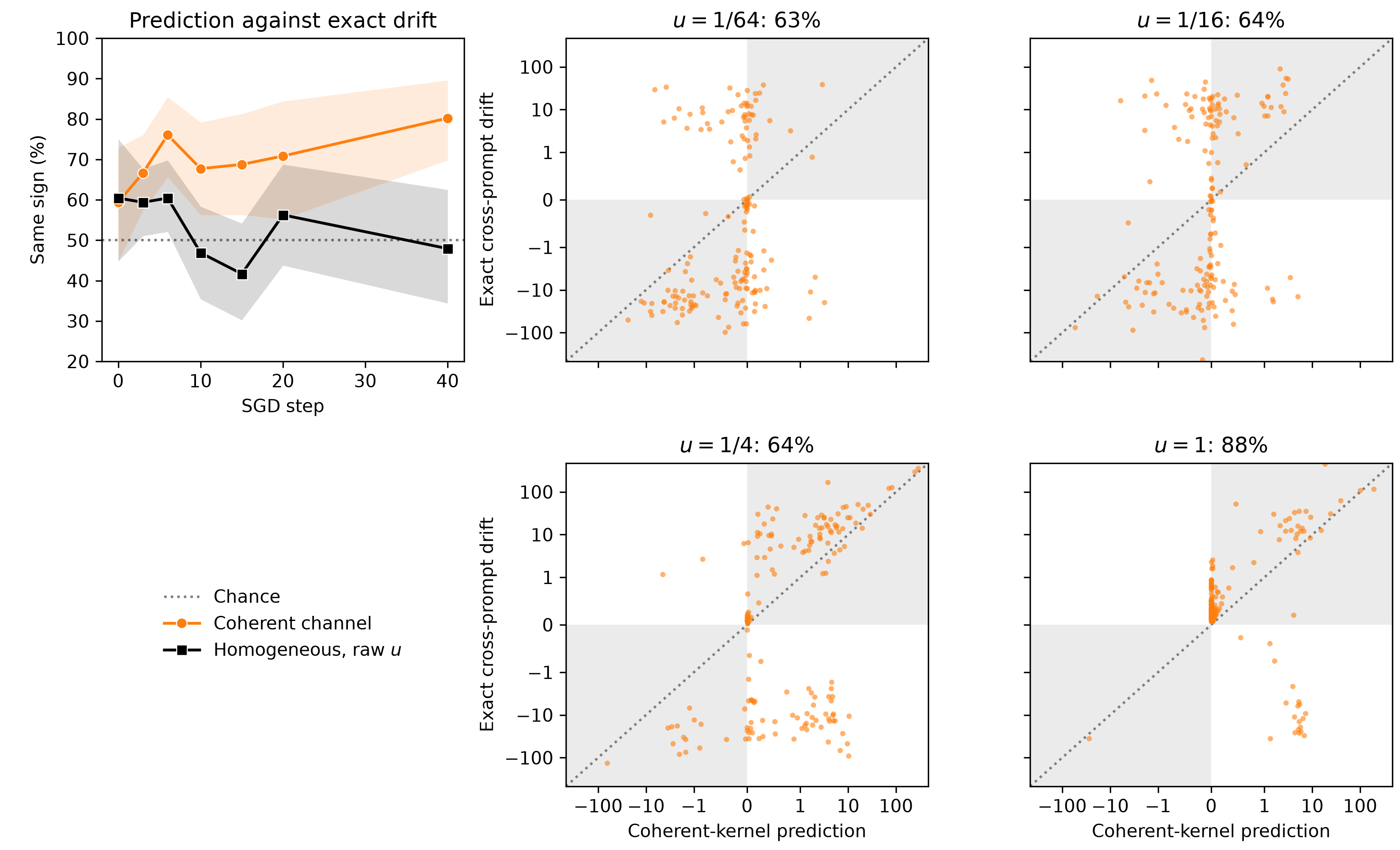}
\caption{The threshold probe. Left: along plain SGD, the fraction of probed prompts at which the coherent-channel prediction, and the homogeneous raw-$u$ formula, have the same sign as the exact cross-prompt log-odds drift, with bootstrap $95\%$ bands over the (seed, universality) runs. Right: the coherent-channel prediction against the exact drift for every probed prompt at every step, one panel per universality, on symmetric log axes. Shaded quadrants are where the two agree in sign, and each title gives the share of points there.}
\label{fig:probe}
\end{figure}

Two questions remain open after Section~\ref{sec:theory}: whether Proposition~\ref{prop:threshold} predicts anything about the sweeps, given that with raw $u$ it predicts suppression below $u\approx\bar V$, which starts at \NuTwoBaseLky{}, and whether the instantaneous prediction can be compared with the exact drift before the verifier pass rate saturates. Figure~\ref{fig:mechanism} (bottom right), Figure~\ref{fig:probe}, and the numbers below answer both.

\paragraph{Where the mass sits.} Equation~\eqref{eq:ueff} has two readings (Figure~\ref{fig:mechanism}, bottom right). The mass-only reading sets every kernel entry to one and asks what share of the policy's mass on $h$ sits on accepting prompts; the kernel-weighted reading includes the entries, and at small $m$ is dominated by the own-prompt entry $\kappa_s$, which is many times the cross-prompt entries. In the headline bases the mass-only reading is $\UeffMassEOneLo{}$, $\UeffMassEOneFour{}$, $\UeffMassEOneSixteen{}$, and $\UeffMassEOneHi{}$ at $u=\nicefrac{1}{64}, \nicefrac{1}{16}, \nicefrac14$, and $1$, and the kernel-weighted reading is $\UeffEOneLo{}$, $\UeffEOneFour{}$, $\UeffEOneSixteen{}$, and $\UeffEOneHi{}$. The supervised stage puts about \RepEOneLoPh{} of mass on an idiosyncratic leak at the one prompt that accepts it and \RepEOneLoPhOut{} elsewhere, so the accepting prompts hold several times their proportional share of the sum. In the blind base the mass on $h$ is nearly uniform over prompts, so the mass-only reading is close to $u$ ($\UeffMassENineLo{}$, $\UeffMassENineFour{}$, $\UeffMassENineSixteen{}$, $\UeffMassENineHi{}$) and the kernel-weighted reading is $\UeffENineLo{}$, $\UeffENineFour{}$, $\UeffENineSixteen{}$, and $\UeffENineHi{}$. At $u=1$ the two bases represent $h$ with similar cross-prompt coherence ($\RepEOneCoh{}$ in the headline base against $\RepENineCoh{}$ in the blind base), and the blind base carries less mass on it (\RepEOnePh{} against \RepENinePh{}) because its contamination budget is split between two families. This is the representation difference that has to be identified before the blind-base controls can be interpreted: at $u=1$ the coherence is similar and the mass is lower, which works against amplification there; at $u<1$ the bases differ in where the mass sits and also in coherence, which is lower in the headline base ($\RepEOneLoCoh{}$ against $\RepENineLoCoh{}$ at $u=\nicefrac{1}{64}$, three seeds). The weights $w_{x'}$ in Equation~\eqref{eq:ueff} are positive for most entries: \ProbeNegKFrac{} of the same-string kernel entries at the headline bases are negative, carrying \ProbeNegWeightShare{} of the absolute weight, so $\ueff$ is a fraction in $[0,1]$ to a good approximation there.

\paragraph{Predicted against exact drift.} Along SGD the verifier pass rate of the universal-leak condition reaches \ProbeVbarHiLast{} by step \ProbeVbarStepLast{}, while the idiosyncratic condition is still at \ProbeVbarLoLast{}, so the threshold rises quickly at high $u$ and slowly at low $u$. Over all probed (condition, step) means the coherent-channel prediction agrees in sign with the exact cross-prompt log-odds drift at \ProbeSignAgreeCond{} of points (\ProbeNCond{} points), and the homogeneous leading-order formula with its $1/N$ correction at \ProbeSignAgreeLeadCond{}; per probed prompt the agreement is \ProbeSignAgreePts{} and \ProbeSignAgreeLeadPts{} over \ProbeNPts{} prompts. The median ratio of the residual to the exact drift is \ProbeResidShareMedian{}. The channel the proposition isolates is therefore real, in the sense that its sign and the sign of the exact cross-prompt drift agree well above chance at the level of condition means and above chance per prompt, while the homogeneous formula is at chance per prompt. But it is not the whole cross-prompt drift: the residual $R$ of Definition~\ref{def:coherent} is of the same order as the channel itself. We present the proposition accordingly, as a statement about one channel and the quantity that sets its sign, and we use its heterogeneous form only to interpret the difference between the headline sweep and the blind-base controls.

\section{Additional Results}
\label{app:extra}

\paragraph{Every learning rate.} Table~\ref{tab:lr} lists, for each training setup, learning rate, and arm, the slope of $\Delta\rho$ per doubling of $m$ with its bootstrap interval, the drift at the two ends of the sweep, the number of seeds in which the idiosyncratic leak is suppressed and the universal leak amplified, and the final verifier reward at $u=1$.

\begin{table}[htbp]
\centering\footnotesize
\scriptsize\setlength{\tabcolsep}{2pt}
\begin{tabular}{llrrrrrrrl}
\toprule
setup & arm & LR & slope & $95\%$ CI & $\Delta\rho(\nicefrac1{64})$ & $\Delta\rho(1)$ & neg.\ at $\nicefrac1{64}$ & pos.\ at $1$ & reward at $1$ \\
\midrule
Adam & shared & $0.0003$ & $+0.0075$ & $[+0.0030, +0.0121]$ & $-0.103$ & $-0.051$ & 10/10 & 2/10 & 0.997 \\
Adam & shared & $0.001$ & $+0.0176$ & $[+0.0086, +0.0296]$ & $-0.102$ & $+0.028$ & 10/10 & 5/10 & 1.000 \\
Adam & shared & $0.003$ & $+0.0996$ & $[+0.0753, +0.1251]$ & $-0.098$ & $+0.600$ & 10/10 & 10/10 & 1.000 \\
Adam & shared & $0.01$ & $+0.1287$ & $[+0.1151, +0.1399]$ & $+0.210$ & $+0.869$ & 0/10 & 10/10 & 1.000 \\
Adam & shared & $0.03$ & $+0.0456$ & $[+0.0246, +0.0632]$ & $+0.331$ & $+0.651$ & 0/10 & 10/10 & 0.319 (diverged) \\
Adam & per-prompt & $0.0003$ & $+0.0000$ & $[-0.0000, +0.0001]$ & $-0.000$ & $-0.000$ & 8/10 & 2/10 & 0.527 \\
Adam & per-prompt & $0.001$ & $+0.0001$ & $[-0.0000, +0.0002]$ & $+0.001$ & $+0.002$ & 2/10 & 10/10 & 0.644 \\
Adam & per-prompt & $0.003$ & $+0.0004$ & $[+0.0000, +0.0009]$ & $+0.020$ & $+0.021$ & 0/10 & 10/10 & 0.870 \\
Adam & per-prompt & $0.01$ & $+0.0020$ & $[+0.0007, +0.0032]$ & $+0.110$ & $+0.116$ & 0/10 & 10/10 & 0.982 \\
Adam & per-prompt & $0.03$ & $+0.0028$ & $[+0.0006, +0.0052]$ & $+0.161$ & $+0.164$ & 0/10 & 10/10 & 0.997 \\
Sampled GRPO & shared & $0.0003$ & $+0.0094$ & $[+0.0047, +0.0145]$ & $-0.102$ & $-0.032$ & 10/10 & 3/10 & 0.996 \\
Sampled GRPO & shared & $0.001$ & $+0.0942$ & $[+0.0743, +0.1135]$ & $-0.098$ & $+0.623$ & 10/10 & 10/10 & 0.998 \\
Sampled GRPO & shared & $0.003$ & $+0.1675$ & $[+0.1594, +0.1742]$ & $-0.077$ & $+0.869$ & 10/10 & 10/10 & 1.000 \\
Sampled GRPO & per-prompt & $0.0003$ & $-0.0000$ & $[-0.0001, +0.0000]$ & $-0.005$ & $-0.005$ & 10/10 & 0/10 & 0.509 \\
Sampled GRPO & per-prompt & $0.001$ & $-0.0001$ & $[-0.0002, +0.0000]$ & $-0.016$ & $-0.016$ & 10/10 & 0/10 & 0.584 \\
Sampled GRPO & per-prompt & $0.003$ & $-0.0002$ & $[-0.0005, +0.0001]$ & $-0.042$ & $-0.041$ & 10/10 & 0/10 & 0.770 \\
Plain SGD & shared & $0.3$ & $+0.1834$ & $[+0.1773, +0.1889]$ & $-0.077$ & $+0.867$ & 10/10 & 10/10 & 0.987 \\
Plain SGD & shared & $1$ & $+0.0982$ & $[+0.0718, +0.1161]$ & $+0.153$ & $+0.771$ & 0/10 & 9/10 & 0.906 \\
Plain SGD & shared & $3$ & $-0.0008$ & $[-0.0197, +0.0158]$ & $+0.274$ & $+0.299$ & 0/10 & 9/10 & 0.119 (diverged) \\
Plain SGD & per-prompt & $0.3$ & $-0.0001$ & $[-0.0003, +0.0000]$ & $-0.017$ & $-0.017$ & 10/10 & 0/10 & 0.534 \\
Plain SGD & per-prompt & $1$ & $-0.0004$ & $[-0.0008, +0.0000]$ & $-0.045$ & $-0.045$ & 10/10 & 0/10 & 0.651 \\
Plain SGD & per-prompt & $3$ & $-0.0006$ & $[-0.0014, -0.0000]$ & $-0.076$ & $-0.077$ & 10/10 & 0/10 & 0.837 \\
\bottomrule
\end{tabular}%

\caption{The universality effect at every learning rate. Slopes are per doubling of $m$; intervals are bootstrap $95\%$ over ten seeds. The last column is the final verifier reward at $u=1$ for that arm, with divergent settings marked; per-prompt rewards below $0.9$ are still rising at step \NSteps{}.}
\label{tab:lr}
\end{table}

\begin{figure}[htbp]
\centering
\includegraphics[width=\linewidth]{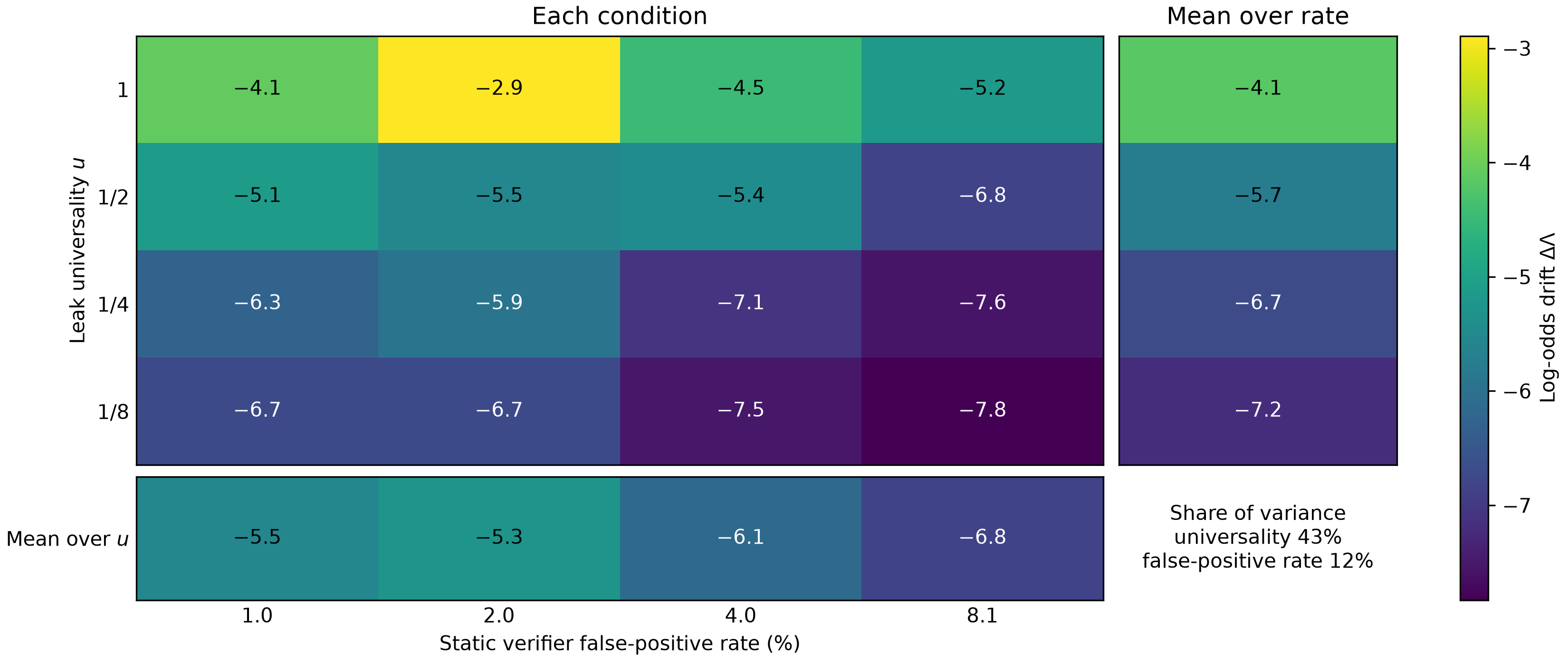}
\caption{The factorial as a field. Each cell is one (universality, false-positive rate) condition; darker is a leak more strongly suppressed relative to the filter baseline. The right column and bottom row are the marginal means on the same color scale. Total base hack mass is fixed at $\varepsilon$ in every cell, so the horizontal trend reflects the same mass spread over more accepted strings rather than a general property of audits.}
\label{fig:field}
\end{figure}

\begin{figure}[htbp]
\centering
\includegraphics[width=\linewidth]{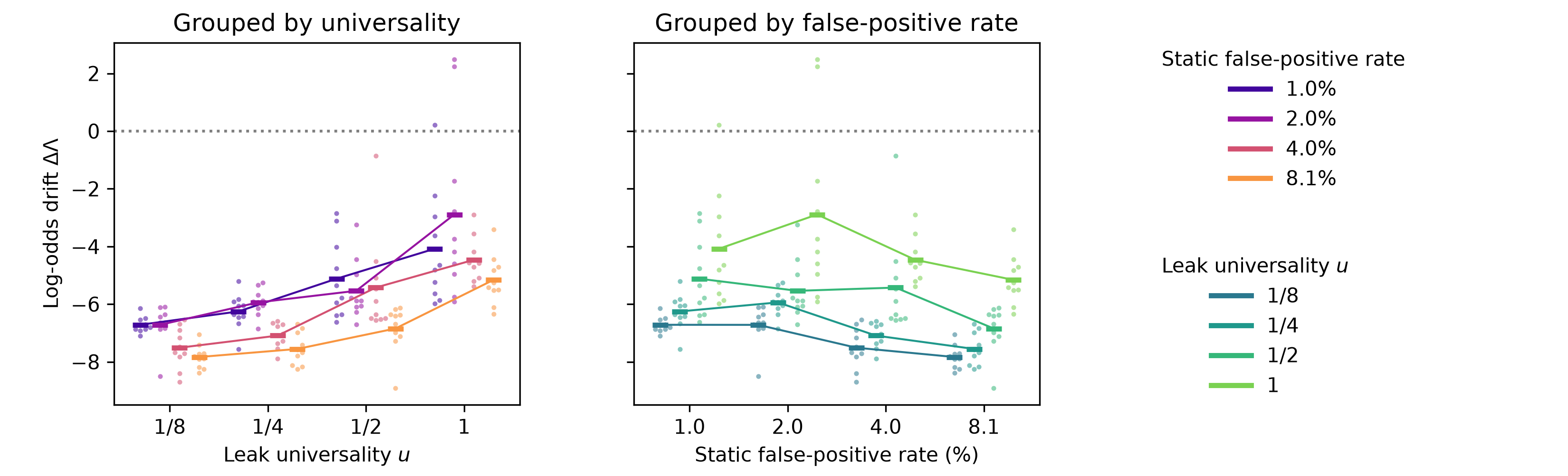}
\caption{The factorial seed by seed, common $y$-scale: dots are seeds, bars cell means, and lines join the means of one color. Left: log-odds drift grouped by universality, one color per false-positive rate. Right: the same data grouped by the static false-positive rate, one color per universality. Varying the accepted wrong responses per prompt from \NuThreeKRange{} moves the false-positive rate over \NuThreeFprRange{} and the drift from $\NuThreeDLLoK{}$ to $\NuThreeDLHiK{}$; the same eightfold range of universality moves it from $\NuThreeDLLoU{}$ to $\NuThreeDLHiU{}$.}
\label{fig:factorial}
\end{figure}

\paragraph{Universality against false-positive rate (E3).} Figure~\ref{fig:field} crosses universality with the static false-positive rate, both swept eightfold, with the total base hack mass held at $\varepsilon$ for every $k$; Figure~\ref{fig:factorial} shows the same data seed by seed on a common scale. A variance decomposition over the \NuThreeNSeeds{}-seed cells attributes \NuThreeEtaM{} of the variance in $\Delta\Lambda$ to universality and \NuThreeEtaK{} to the false-positive rate. Averaged over $k$, $\Delta\Lambda$ runs from $\NuThreeDLLoU{}$ to $\NuThreeDLHiU{}$; averaged over $u$, only from $\NuThreeDLLoK{}$ to $\NuThreeDLHiK{}$. The mild decrease with $k$ is a property of this design, in which accepting more strings spreads a fixed base mass thinner (Appendix~\ref{app:details}), and we do not read it as a general prediction about audits. The robust content is the comparison of magnitudes: the false-positive rate is the weaker variable.

\paragraph{Prompt-set size (E7).} At fixed $u$ the leading term of the homogeneous form of \eqref{eq:threshold} carries no dependence on the number of prompts; the $1/N$ remainder differs fourfold between $N=16$ and $N=64$. Repeating the sweep at both sizes (\NuFourNSeeds{} seeds) gives, at $u=1$, $\NuFourUOneSmall{}\pm\NuFourUOneSmallSem{}$ against $\NuFourUOneBig{}\pm\NuFourUOneBigSem{}$, whereas at the same block size $m=\NuFourSameM{}$ (so $u=1$ for the small set and $u=\nicefrac14$ for the large one) they sit at $\NuFourUOneSmall{}$ against $\NuFourSameMBig{}$. Across the \NuFourNShared{} matched values of $u$ the two sizes agree to within \NuFourMatchedGapLo{} at the three smallest values and differ by at most \NuFourMatchedGap{}, at $u=1$, where the seed spread is widest. Within the precision of ten seeds, what matters is the fraction $u$ rather than the block size $m$.

\begin{figure}[htbp]
\centering
\includegraphics[width=\linewidth]{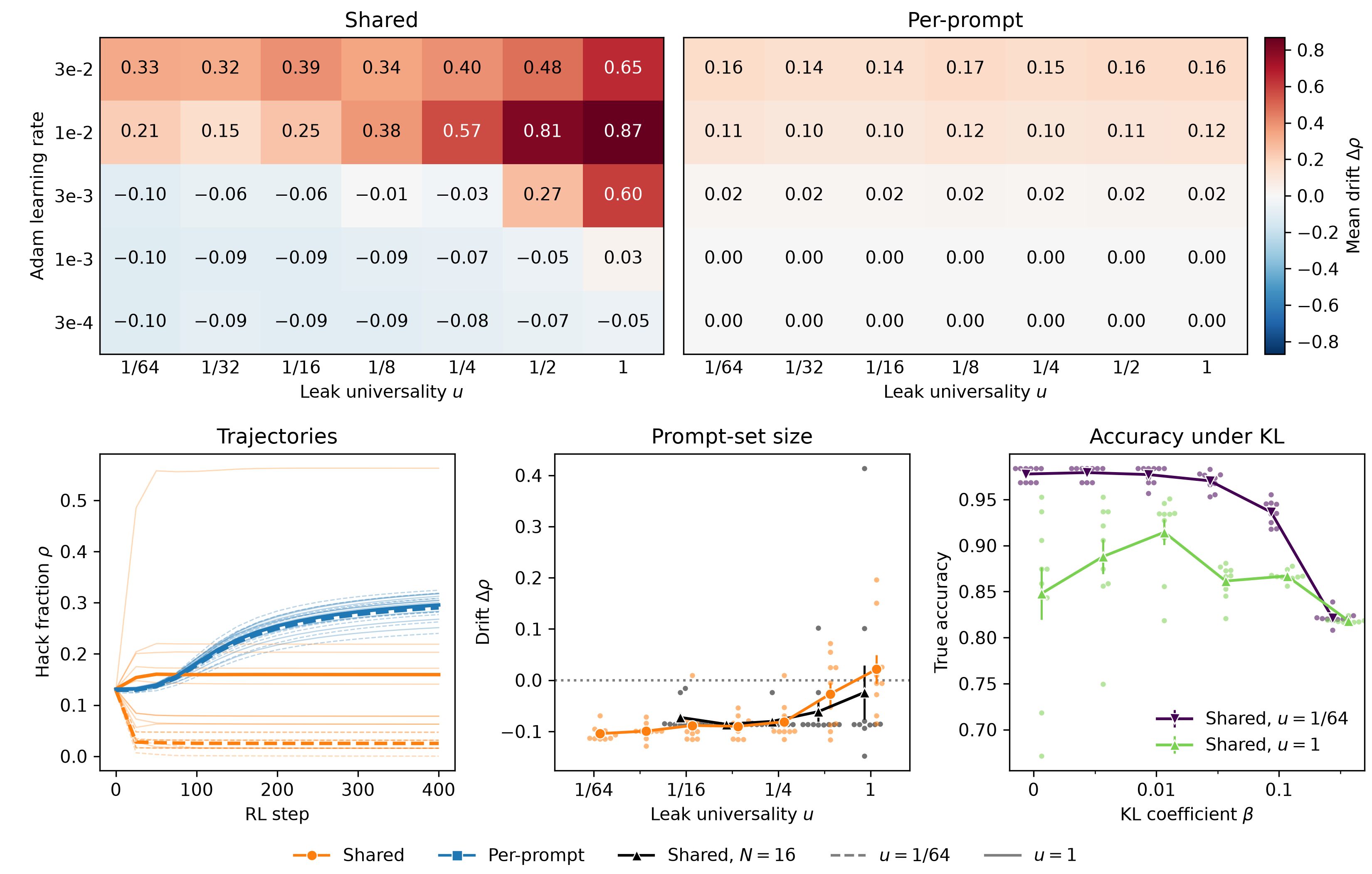}
\caption{Top: the universality effect across the whole Adam learning-rate grid for both arms, as the mean drift over ten seeds on one color scale. Bottom left: hack-fraction trajectories for the extreme universality conditions, both arms, with thin lines for seeds; the per-prompt curves coincide. Bottom center: the sweep repeated at two prompt-set sizes, plotted against $u$, with $N=64$ in orange. Bottom right: true accuracy at the two ends of the sweep as the KL coefficient grows. In the bottom center and right panels small dots are seeds and markers their means $\pm1$ s.e.m.}
\label{fig:appendix}
\end{figure}

Figure~\ref{fig:appendix} collects the remaining robustness views: the universality effect across the whole Adam grid, hack-fraction trajectories at the two extreme conditions for both arms, the sweep at two prompt-set sizes, and true accuracy under the KL penalty of Section~\ref{sec:practice}.

\section{Reproducibility}
\label{app:repro}

Every experiment runs on CPU; no GPU is used. The code, the raw result files, and the figure code that regenerates every figure and every number quoted in the text are released together at \url{https://github.com/Rohan-Pandey1729/RLVR-Reward-Hacking}. The macro file that holds the numbers is written by the figure code, so nothing is transcribed by hand except settings (the step count, the contamination rate, the GRPO group size) and a few diagnostics quoted inline (the clipping rate and the reproduction tolerances below).

Three levels of reproduction apply. From an archived base checkpoint, on the same machine and code revision, the RL stage is bit-reproducible in weights and in every reported metric. Regenerating a base by rerunning the supervised stage is deterministic, but a base rebuilt under a different version of the numerical libraries differs from an archived one by up to about $2\times10^{-2}$ on the base metrics. A run from scratch therefore reproduces the reported drifts, most of which are several times larger, and the conclusions drawn from them, but not every archived value digit for digit.

E1, E4, E8, E9, E9b, and E10 are archived runs. E2, E3, E5, E6, E7, and E11 were run from base models regenerated with the released code, E2 at the three smallest learning rates of the Adam grid. Seeds $0$--$4$ of E1 come from an earlier code revision whose configuration records carry fewer logging fields, with no training setting differing; their stored base models are bit-identical and the current code reproduces that half to $10^{-8}$. The $m=1$ condition of E1 ran past the reported seed budget during an attempt to raise the budget for every condition; the figure code caps every E1 load at ten seeds, uniformly and independently of results, so the design is balanced and the excess runs enter no reported statistic.

\section{Additional Related Work}
\label{app:related}

\paragraph{Systematic error and amplification.} \citet{khalifa2026countdown} showed that about $1\%$ of hacking demonstrations in supervised fine-tuning primes hacking under RLVR, and \citet{zhao2025echo} showed that RL post-training amplifies behaviors already present in pretraining; \citet{zhang2026leaky} reports inheritance while \citet{khalifa2026countdown} and \citet{nishimura2026specgaming} report amplification and specification gaming. Work on what corrupted rewards can do \citep{everitt2017corrupted, skalse2022defining, gao2023scaling} and on group-relative advantage estimators \citep{yang2026biased, zeng2025shrinkage} is orthogonal.

\paragraph{Coherent gradients.} The mutual reinforcement of gradients from examples that agree is the coherent-gradients hypothesis of \citet{chatterjee2020coherent}.

\paragraph{RLVR as filtering.} The KL-regularized optimum as an exponential tilt of the reference model is well established \citep{korbak2022bayesian, gai2025smoothing, kruszewski2026filtering, dymetman2026binary, lochab2026ucpo, rafailov2023dpo, yue2025rlincentivize}. That literature reads the invariance as a statement about diversity among correct answers; we read it as a constraint on the composition of the accepted set and use it as our null model, with a per-prompt policy as a control, which under gradient flow shows the rich-get-richer behavior of softmax policy gradient among tied responses \citep{mei2020global, lochab2026ucpo}.

\paragraph{Verifier error.} \citet{wang2020perturbed} and \citet{cai2025noisyverifier} show that correcting for reward noise helps, not that uncorrected training is robust. \citet{plesner2026imperfect} measure noise tolerance directly, \citet{rad2026rateorfate} map a phase transition in the Youden index, and spurious rewards occasionally help particular models \citep{shao2025spurious}. Audits of verifier false positives \citep{zhao2025onetoken, huang2025verifiers, xu2025tinyv, yuan2026miracle} report gross error rates and leave the alignment of the errors across prompts unmeasured.

\end{document}